%% file: Time-aware_Dynamic_Graph_Embedding.tex
\documentclass[10pt,journal,compsoc]{IEEEtran}
\ifCLASSOPTIONcompsoc
\usepackage[nocompress]{cite}
\else
\usepackage{cite}
\fi
\ifCLASSINFOpdf
\usepackage[pdftex]{graphicx}
\else
\usepackage[dvips]{graphicx}
\fi
\usepackage{booktabs} 
\usepackage{amssymb}
\usepackage{amsfonts}
\usepackage{multirow}
\usepackage{ntheorem}
\usepackage{textcomp}
\usepackage{xcolor}
\usepackage{subfigure}
\usepackage{stfloats}
\usepackage[switch]{lineno}

\usepackage{amsmath}
\usepackage{algorithmic}

\usepackage{array}
\usepackage{url}

\newtheorem{lemma}{Lemma}
\newtheorem*{proof}{Proof:}
\newtheorem{definition}{Definition}

\begin{document}

\title{Time-aware Dynamic Graph Embedding for Asynchronous Structural Evolution}
%
%
%
%

\author{Yu~Yang,
	Hongzhi Yin,
	Jiannong~Cao,~\IEEEmembership{Fellow,~IEEE,}
	Tong Chen,
	Quoc Viet Hung Nguyen,\\
	Xiaofang Zhou,~\IEEEmembership{Fellow,~IEEE,}
	Lei Chen,~\IEEEmembership{Fellow,~IEEE}
	\IEEEcompsocitemizethanks{\IEEEcompsocthanksitem Y. Yang and J. Cao are with the Department
		of Computing, The Hong Kong Polytechnic University, China.\protect\\
		E-mail: \{cs-yu.yang, jiannong.cao\}@polyu.edu.hk  
		\IEEEcompsocthanksitem H. Yin and T. Chen are with the School of Information Technology and Electrical Engineering, The University of Queensland, Australia. \protect\\
		E-mail: \{tong.chen, h.yin1\}@uq.edu.au
		\IEEEcompsocthanksitem Q. V. H. Nguyen is with The Griffith University, Australia. \protect\\
		E-mail: quocviethung.nguyen@griffith.edu.au
		\IEEEcompsocthanksitem X. Zhou and L. Chen are with the Department of Computer Science and Engineering, The Hong Kong University of Science and Technology, China. \protect
		E-mail: \{zxf, leichen\}@cse.ust.hk
	}
}

\markboth{IEEE Transactions on Knowledge and Data Engineering~2023}%
{Shell \MakeLowercase{\textit{et al.}}: Bare Demo of IEEEtran.cls for Computer Society Journals}

\IEEEtitleabstractindextext{%
\begin{abstract}
	Dynamic graphs refer to graphs whose structure dynamically changes over time. 
	Despite the benefits of learning vertex representations (i.e., embeddings) for dynamic graphs, existing works merely view a dynamic graph as a sequence of changes within the vertex connections, neglecting the crucial asynchronous nature of such dynamics where the evolution of each local structure starts at different times and lasts for various durations. To maintain asynchronous structural evolutions within the graph, we innovatively formulate dynamic graphs as temporal edge sequences associated with joining time of vertices (ToV) and timespan of edges (ToE). Then, a time-aware Transformer is proposed to embed vertices' dynamic connections and ToEs into the learned vertex representations. Meanwhile, we treat each edge sequence as a whole and embed its ToV of the first vertex to further encode the time-sensitive information. Extensive evaluations on several datasets show that our approach outperforms the state-of-the-art in a wide range of graph mining tasks. At the same time, it is very efficient and scalable for embedding large-scale dynamic graphs.
\end{abstract}

}

\maketitle

\IEEEdisplaynontitleabstractindextext

%
\IEEEpeerreviewmaketitle

\input{body}
\section*{Acknowledgment}
The work is supported by Shenzhen-Hong Kong-Macau Technology Research Programme Type C (No. SGDX202011 03095203029), Hong Kong Research Grants Council under Theme-based Research Scheme (No. T41-603/20-R), Research Impact Fund (No. R5034-18), Collaborative Research Fund (No. C5026-18G), and General Research Fund (No. 16202722), Natural Science Foundation of China (No. 62072125), PolyU Research and Innovation Office (No. BD4A), the Australian Research Council under ARC Future Fellowship (No. FT210100624), Discovery Project (No. DP190101985), Discovery Early Career Research Award (No. DE200101465, DE230101033), Industrial Transformation Training Centre (No. IC200100022), and Centre of Excellence (No. CE200100025), and was partially conducted in Research Institute for Artificial Intelligence of Things at PolyU and the JC STEM Lab of Data Science Foundations funded by The Hong Kong Jockey Club Charities Trust.

\bibliographystyle{IEEEtran}
\bibliography{bibliography}

\vspace{-35pt}
\begin{IEEEbiography}[{\includegraphics[width=1in,height=1.25in,clip]{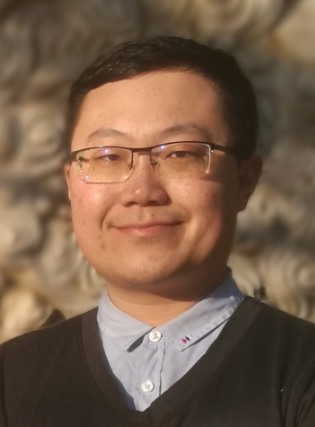}}]{Yu Yang}
	is currently a Research Assistant Professor with the Department of Computing, The Hong Kong Polytechnic University. He received the Ph.D. degree in Computer Science from The Hong Kong Polytechnic University in 2021. His research interests include spatiotemporal data analysis, representation learning, urban computing, and learning analytics.
\end{IEEEbiography}
\vspace{-45pt}
\begin{IEEEbiography}[{\includegraphics[width=1in,height=1.25in,clip]{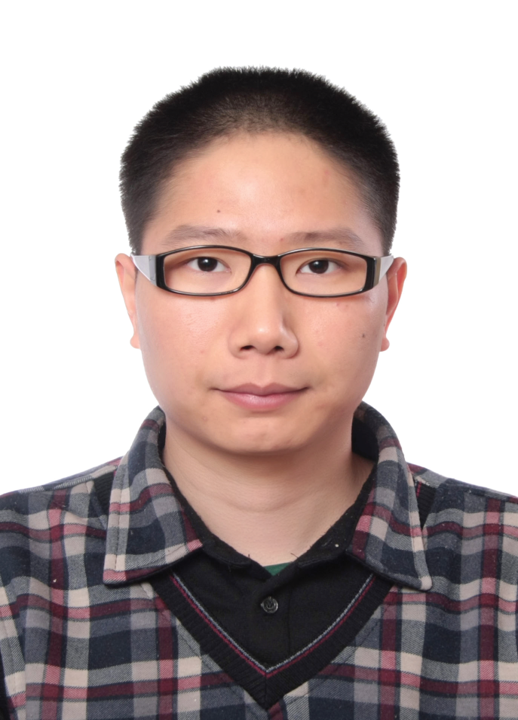}}]{Hongzhi Yin}
	received the Ph.D. degree in computer science from Peking University in 2014. He is an Associate Professor with the University of Queensland. He received the Australia Research Council Discovery Early-Career Researcher Award, in 2015. His research interests include recommendation system, user profiling, topic models, deep learning, social media mining, and location-based services.
\end{IEEEbiography}
\vspace{-45pt}
\begin{IEEEbiography}[{\includegraphics[width=1in,height=1.25in,clip]{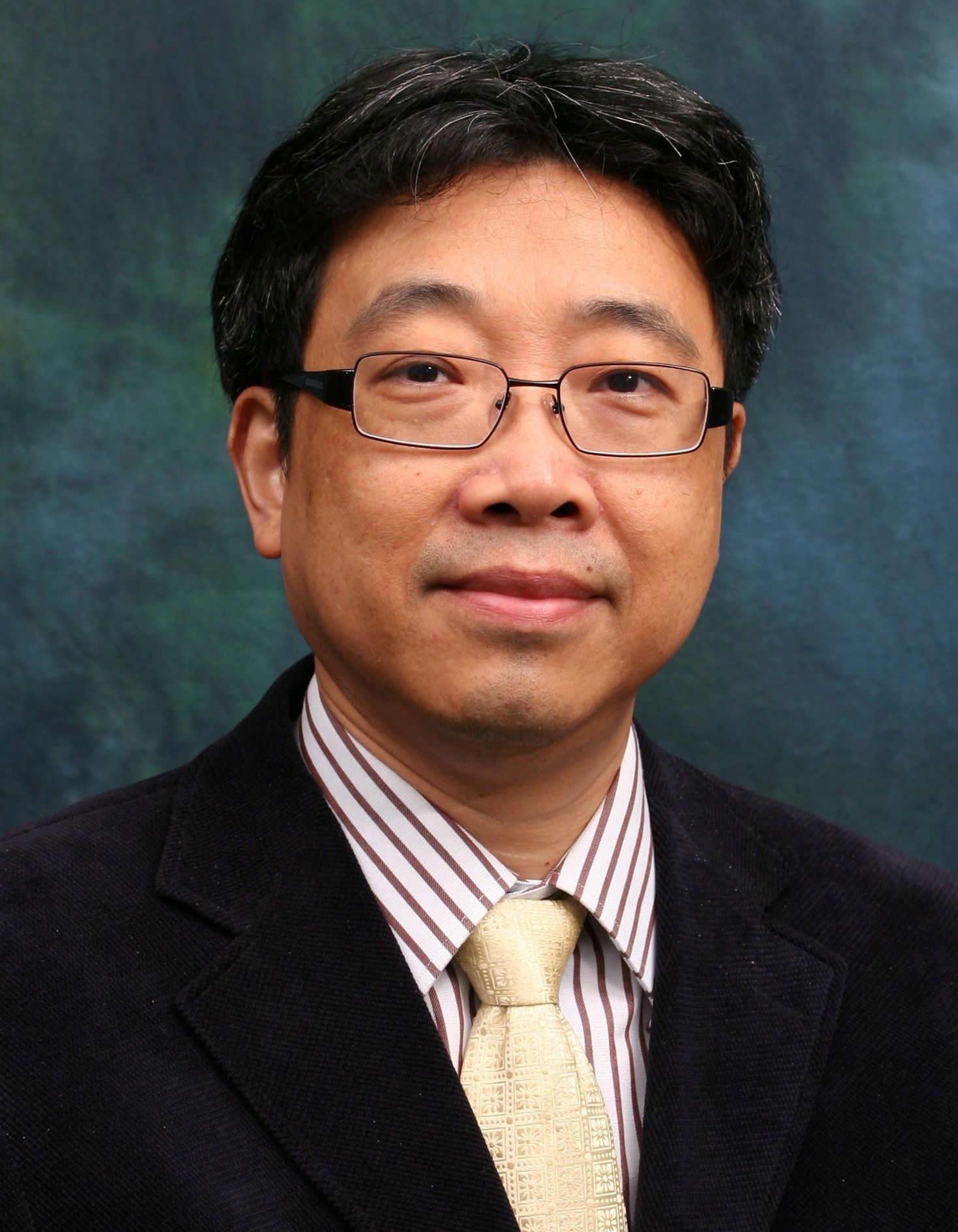}}]{Jiannong Cao}
	(M'93-SM'05-F'15) received the B.Sc. degree in Computer Science from Nanjing University,
	China, in 1982, and the M.Sc. and Ph.D. degrees in Computer Science from Washington State University, USA, in 1986 and 1990, respectively. He is the Chair Professor with the Department of Computing, The Hong Kong Polytechnic University, Hong Kong. His current research interests include big data analytics, edge intelligence, and mobile computing.
\end{IEEEbiography}
\vspace{-45pt}
\begin{IEEEbiography}[{\includegraphics[width=1in,height=1.25in,clip]{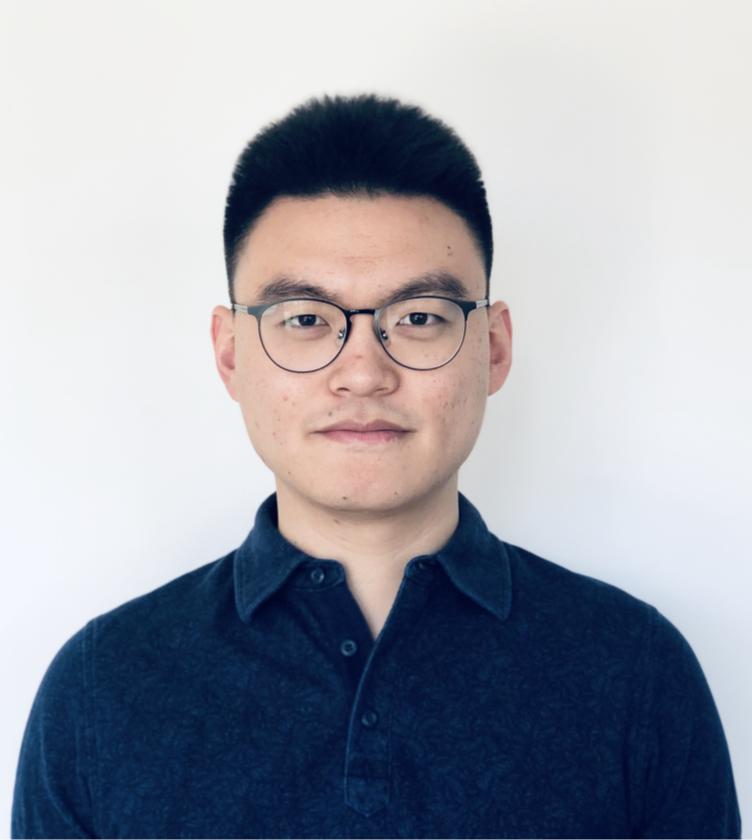}}]{Tong Chen}
	received the Ph.D. degree in computer science from the University of Queensland in 2020. He is a lecturer with Data Science Group, School of ITEE, UQ. His research work has been published on top venues like SIGKDD, ICDE, WWW, IJCAI, AAAI, TKDE, etc., where his research interests include data mining, recommender systems, and predictive analytics.
\end{IEEEbiography}
\vspace{-45pt}
\begin{IEEEbiography}[{\includegraphics[width=1in,height=1.25in,clip]{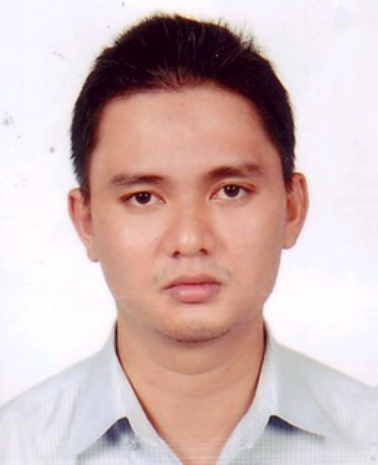}}]{Nguyen Quoc Viet Hung}
	is a Senior Lecturer in Griffith University. He earned his Master and Ph.D. degrees from EPFL (Switzerland). He received the Australia Research Council Discovery Early-Career Researcher Award, in 2020. His research focuses on Data Integration, Information Retrieval, Trust Management, Recommender Systems, etc., with special emphasis on web data, social data and sensor data.
\end{IEEEbiography}
\vspace{-45pt}
\begin{IEEEbiography}[{\includegraphics[width=1in,height=1.25in,clip]{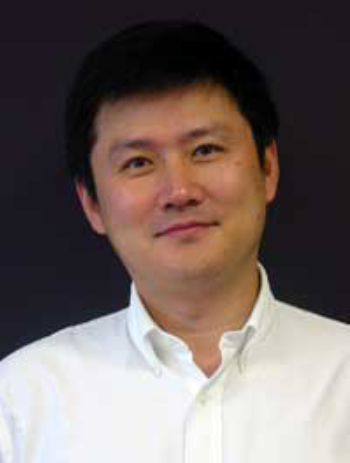}}]{Xiaofang Zhou}
	is a Chair Professor with the Department of Computer Science and Engineering at Hong Kong University of Science and Technology. He received the Ph.D. degree in Computer Science from University of Queensland in 1994. His research focus is to find effective and efficient solutions for analysing large-scale data for business, scientific and personal applications.
\end{IEEEbiography}
\vspace{-45pt}
\begin{IEEEbiography}[{\includegraphics[width=1in,height=1.25in,clip]{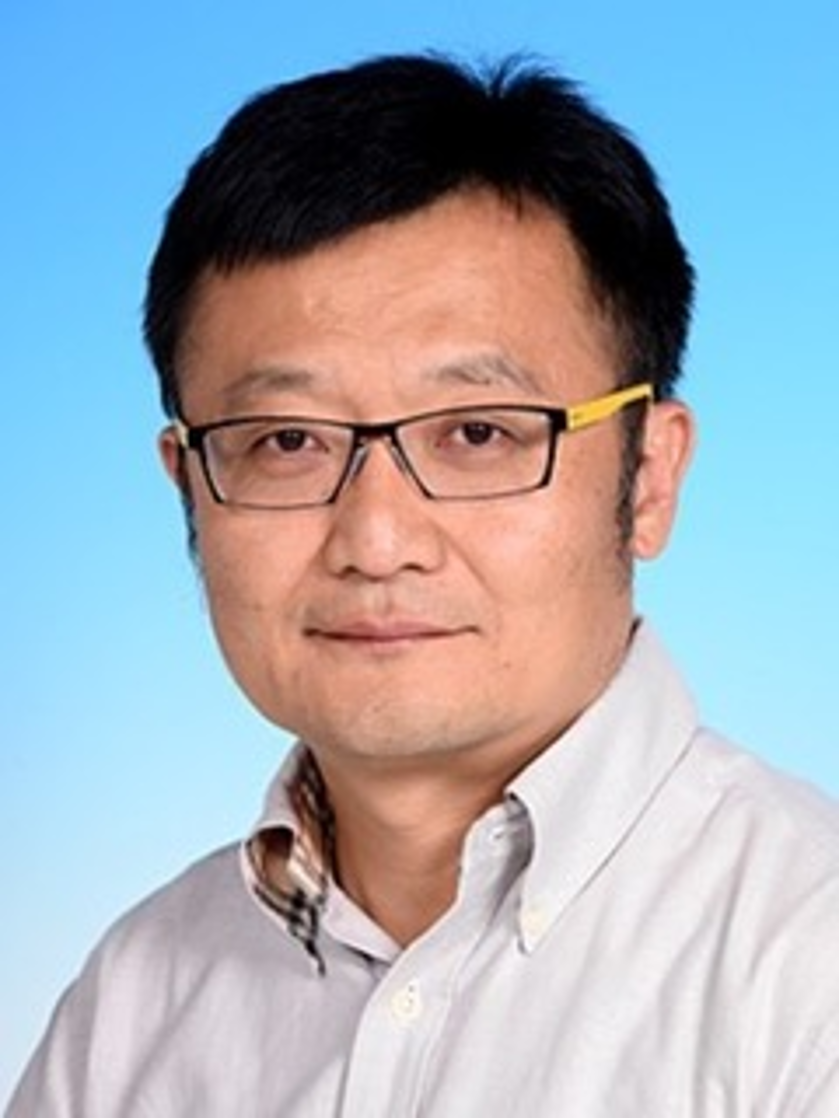}}]{Lei Chen}
	received the Ph.D. degree in Computer Science from the University of Waterloo, Canada, in 2005. He is a Chair Professor with the Department of Computer Science and Engineering at Hong Kong University of Science and Technology. His research interests include data-driven machine learning, graph, and crowdsourcing.
\end{IEEEbiography}

\end{document}

%% file: body.tex
\IEEEraisesectionheading{\section{Introduction}\label{sec:Introduction}}

\IEEEPARstart{G}{raphs} are one of the most widely used data structures to represent pairwise relations between entities.
In many real-world applications, these relations naturally change over time, making the graph dynamic.
For example, in an online forum, users can post messages at any time and form a discussion network.
Some of them join the discussion by replying or citing the published posts, thus forming edges among existing vertices. Some may choose to unfollow other users or become inactive in the discussion, and this will make them disconnected from other vertices. 
Since vertices can join, leave and re-join a network at any time, the dynamics of graphs are beneficial for modeling various types of data like traffic, financial transactions, and social media.

Dynamic graph embedding is an effective means to encode the evolution of vertices' connections and properties into vector representations to facilitate downstream applications. In general, to capture the dynamic changes of vertex connections over time, existing approaches represent the dynamic graph as either a snapshot graph sequence (SGS)~\cite{zhu2016scalable}\cite{goyal2019dyngraph2vec}\cite{zhu2018high}\cite{zhou2018dynamic}\cite{velickovic2018graph}\cite{manessi2020dynamic}\cite{xiong2019dyngraphgan} or neighborhood formation sequences (NFS)~\cite{nguyen2018continuous}\cite{fathy2020temporalgat}\cite{chang2020continuous}\cite{zuo2018embedding}. On one hand, SGS segments the graph information into several time slots, where a static snapshot graph is built to represent the node connectivity within each time slot. On the other hand, NFS captures dynamic graph information by using temporal random walk to sample the sequential connections among vertices. 

\begin{figure}
	\centerline{\includegraphics[width=0.45\textwidth]{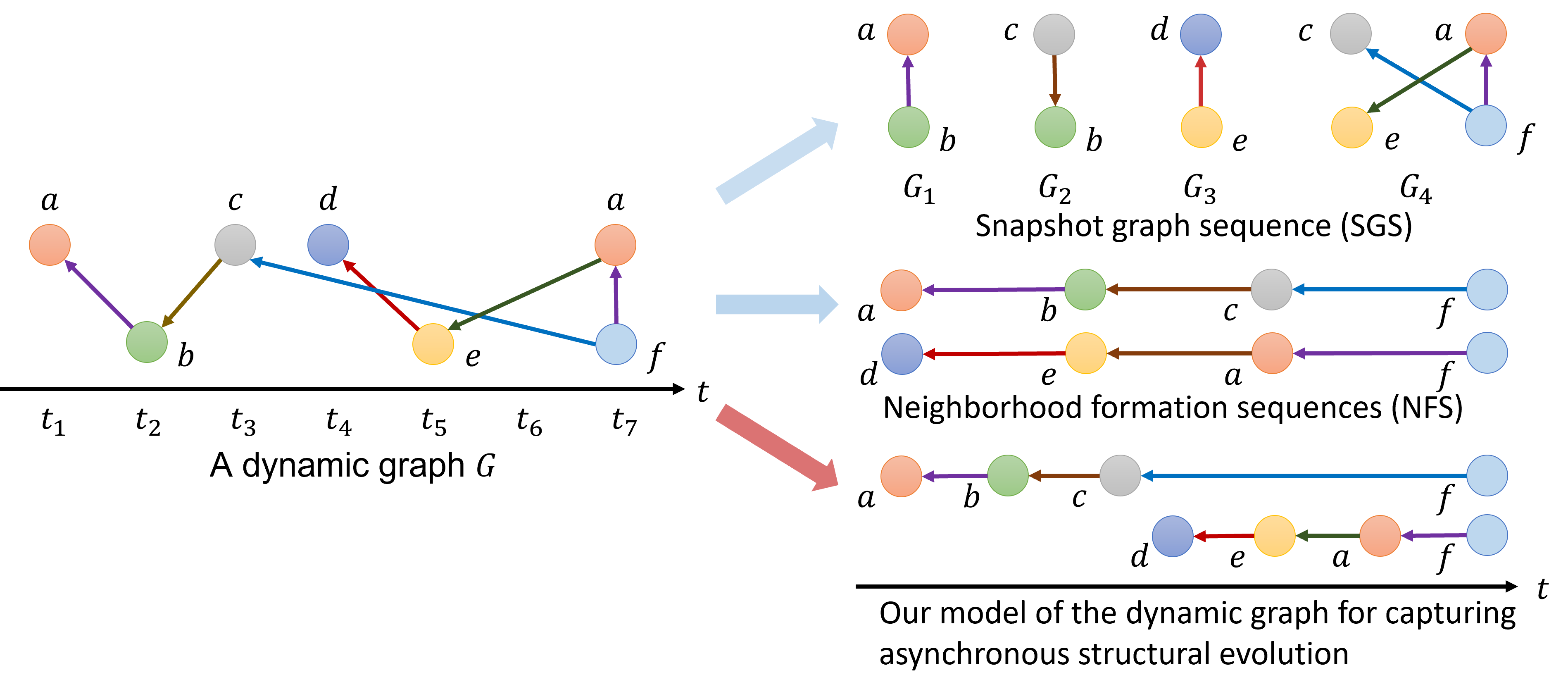}}
	\caption{Modeling a dynamic graph with SGS, NFS, and our approach.}
	\label{fig:dynamic_graph_modeling}
\end{figure}

Unfortunately, both SGS and NFS incur inevitable information loss about the dynamics properties of a graph, thus impeding the expressiveness of the eventually learned node embeddings. In Fig. \ref{fig:dynamic_graph_modeling}, we provide an example on how SGS and NFS represent the dynamic graph $G$, respectively. For SGS, a snapshot graph is built for each of the 4 time slots ($[t_3,t_4]$ and $[t_5,t_6]$ are omitted as there are no new connections). Obviously, SGS tends to suffer from sparsity issues with a fine-grained time granularity, e.g., apart from $G_4$, the vertices and edge in $G_1$, $G_2$, and $G_3$ are too sparse for effectively learning node embeddings. Should these three snapshot graphs be merged to alleviate the sparsity, the subtle structural evolutions at different time steps will be neglected. Furthermore, a largely overlooked fact in dynamic graphs is that, the structural evolutions are \textbf{asynchronous}, i.e., the exact time consumed by each edge update varies in different cases. For example, on the left of Fig. \ref{fig:dynamic_graph_modeling}, the connections among vertices $\{a,b,c,f\}$ and $\{d,e,a,f\}$, evolve asynchronously since the edges have different starting times and durations. As a result, modeling $G$ as a SGS shown in Fig. \ref{fig:dynamic_graph_modeling} will interrupt the continuous evolution of vertices $\{a,b,c,f\}$ with an irrelevant snapshot $G_3$. Therefore, incorrect graph dynamics will be embedded, as such asynchronous temporal information is totally lost when treating a dynamic graph as an array of static snapshots.

Compared with SGS, NFS is more robust to sparsity as a certain amount of substructures of the dynamic graph are considered for different time steps owing to the temporal random walk strategy. However, it still fails to capture the graph's asynchronous structural evolutions.
Taking $G$ in Fig. \ref{fig:dynamic_graph_modeling} as an example, the local structure among $\{a,b,c,f\}$ and $\{d,e,a,f\}$ starts evolving at $t_1$ and $t_4$, respectively.
The former structure takes $6$ time steps to completely link four vertices, which is slower than the latter one that takes just $3$ steps. Nevertheless, as shown in the middle-right part of Fig. \ref{fig:dynamic_graph_modeling}, the NFS model neglects this important temporal discrepancy, thus being unable to preserve such asynchronous structural evolutions.

In this paper, we study the important yet overlooked problem in dynamic graph embedding, namely fully capturing the asynchronous structural evolutions of a graph. Such asynchronous nature can be interpreted as the starting time and duration of a new edge that emerges between two vertices, and is of great importance to a wide range of time-sensitive applications, e.g., online information propagation modeling and crime prediction. 
However, to achieve the goal, we are confronted with three major challenges.
(1) Capturing the asynchronous structural evolutions in a dynamic graph.
The representation of a dynamic graph should capture not only the dynamic connections among vertices but also the starting time and duration of such evolutions, so as to provide sufficient knowledge about the asynchronous characteristics of a dynamic graph.
(2) Embedding spatial and temporal dynamics of edges.
When vertices join, leave, and re-join the graph, the edges they formed will change accordingly, which brings spatial dynamics. Meanwhile, it takes a different amount of time for vertices to establish new links, leading to dramatic temporal dynamics. To preserve the dynamic edge formation, both the spatial and temporal dynamics should be fully encoded in the dynamic node embeddings.
(3) Preserving asynchronous evolutions of local structures.
Some local structures evolve early, while others evolve later. The embedding algorithm should account for the evolution starting time for every local structure along with its dynamic edges, such that the patterns within asynchronous structural evolutions can be effectively learned to facilitate predictions into the future.

In light of these challenges, we propose to represent a dynamic graph as a set of temporal edges, coupled with the respective joining time of vertices (ToV) and timespan of edges (ToE) as two crucial indicators of the asynchronous properties.
A time-centrality-biased temporal random walk is then developed to sample the local structures as temporal edge sequences. The ToV of the first vertex in a temporal edge sequence corresponds to the evolution starting time of the local structure, and the total ToE indicates how long the evolution of the current local structure takes. Hence, the temporal edge sequences successfully capture the asynchronous structural evolutions for learning expressive node embeddings. Moreover, we propose a novel \textbf{T}ime-\textbf{A}ware \textbf{D}ynamic \textbf{G}raph \textbf{E}mbedding (TADGE) algorithm to embed the asynchronous structural evolutions into vertex representations.
In order to thoroughly incorporate both the spatial and temporal dynamics of the edge formation, we design a time-aware Transformer model. Intuitively, a vertex forms a new edge by linking another vertex that has high affinity with it.
Therefore, we build a Transformer \cite{vaswani2017attention} to embed vertices' dynamic connections as their self-attentive embeddings through an encoder-decoder framework, making the learned embeddings for every vertex carry the structural information and ToE from its connected neighbors. In order to preserve the asynchronous evolution of local structures, we learn an overall representation of every edge sequence by accounting for its evolution starting time.
Lastly, we fuse the vertex- and sequence-level representations to generate the final embedding for each vertex, encoding the patterns of its asynchronous structural evolutions in the dynamic graph to support downstream tasks.

Our contributions are highlighted as follows:
\begin{itemize}
	\item \textbf{A New Problem.}
	To the best of our knowledge, we are the first to study the problem of embedding the asynchronous structural evolutions of a dynamic graph, in which the evolution starting time and duration of each local structures vary significantly.
	\item \textbf{A Novel Representation of Dynamic Graphs.}
	We propose a time-centrality-biased temporal random walk to innovatively formulate the dynamic graph as temporal edge sequences associated with ToV and ToE tags, which preserves the asynchronous structural evolutions for learning vertex embeddings.
	\item \textbf{A New Approach for Dynamic Graph Embedding.}
	We propose TADGE, which is a novel time-aware graph embedding approach that learns expressive dynamic vertex embeddings by fusing information of both the dynamic edge formation and the asynchronous local structure evolutions.
	\item \textbf{Extensive Experiments.}
	We conduct experiments on several large-scale public datasets on dynamic graphs. Experimental results demonstrate the superiority of TADGE, which outperforms the state-of-the-arts and also shows significant advances in training efficiency and scalability.
\end{itemize}

The remainder of this paper is organized as follows.
We present the problem definition in Section 2, followed by our proposed time-centrality-biased temporal random walk and TADGE algorithm in Sections 3 and 4.
Experimental results are reported in Section 5, followed by the literature review in Section 6 before we conclude the paper.

\section{Problem Definition}

In this section, we give the definition of a dynamic graph and formulate the problem of dynamic graph embedding for preserving the asynchronous structural evolutions.

In a dynamic graph $G$, vertices join the graph at any time as either isolated vertices or forming edges with existing ones.
Thus, we denote a vertex $v_i$ joining $G$ at time $t$ as $v_i^t$ where $t$ is the joining time of the vertex (ToV) and $i=1,2,\cdots,n$.
Since vertices join, leave and rejoin the graph dynamically, which triggers the structural evolution, we denote all ToVs of $v_i$ as its ToV set $T_{v_i}$ in which the number of appearances of $v_i$ in $G$ is $|T_{v_i}|$.
$|\cdot|$ is an operator for counting the number of elements in a set.
When $v_i^t$ links to an existing vertex $v_j^{t'}$ whose ToV $t'<t$ and $j=1,2,\cdots,n$, they form a temporal edge $e_{v_i,v_j}^{t, \delta}=(v_i,v_j,t,\delta,w)$ at time $t$, where $\delta=t-t'$ is the timespan of the edge (ToE) indicating how long it takes for $v_i^t$ to form an edge with $v_j^{t'}$, and $w$ is the edge weight.
If $i=j$, a self-link is formed.

Next, we give a formal definition of a dynamic graph that is composed of the dynamic appearing of vertices at different times with the edges they formed, and then we formulate the problem of dynamic graph embedding for preserving asynchronous structural evolutions.

\begin{definition}
	\label{Def:Dynamic_graph}
	\textbf{Dynamic Graph.} A dynamic graph $G=\{V,E,T_V\}$, where $V=\{v_1,v_2,\cdots\}$ denotes a vertex set containing $|V|$ vertices in $G$, and $T_V=\{T_{v_1},T_{v_2},\cdots\}$ is the ToV set of every vertex in $V$.
	$E$ is the temporal edge set in which $e_{v_i,v_j}^{t, \delta}=(v_i,v_j,t,\delta,w)\in E$ is a temporal edge linking $v_i,v_j\in V$.
	$t\in T_{v_i}$ is the ToV of an upcoming vertex $v_i$ and indicates the edge forming time.
	$t'\in T_{v_j}$ is the ToV of an existing vertex $v_j$ and $t'<t$.
	$\delta=t-t'$ is the ToE.
	$w$ is the edge weight.
\end{definition}

\begin{definition}
	\label{Def:Dynamic_graph_embedding}
	\textbf{Dynamic Graph Embedding.} Given a dynamic graph $G=\{V,E,T_V\}$, the objective is to learn a mapping function $f:v\mapsto r_v \in \mathbb{R}^k$ for $\forall v\in V$ such that the representation $r_v$ preserves the asynchronous structural evolution of $v$ in terms of asynchronous evolution starting time and duration, where $k$ is a positive integer indicating the dimension of $r_v$.
\end{definition}

Since we aim to embed the asynchronous structural evolutions in a dynamic graph, we assume $|V|$ is well-known and mainly focus on the edge updates in our study. Meanwhile, our approach is compatible with any inductive learning schemes to handle newly joined vertices.

\section{Capturing Asynchronous Structural Evolutions in the Dynamic Graph}
\label{sec:temporal_walk}

Directly learning embeddings from the origin graph $G$ is difficult due to its complexity and dynamics.
Thus, it is necessary to transform the original dynamic graph into a proper format that captures its dynamic structural evolution and is easy for later embedding.
However, regardless of modeling the dynamic graph as either SGS or NFS, the asynchronous structural evolutions cannot be captured fairly well.
Inspired by \cite{nguyen2018continuous} and \cite{chen2019exploiting}, we propose a time-centrality-biased temporal random walk to sample the dynamic graph and model it as a set of temporal edge sequences that properly capture the asynchronous structural evolutions.

Specifically, we first randomly select a set of initial vertices via uniform distribution, which appear at $m$ different times and satisfy $|V|\le m< \sum_{v\in V}|T_{v}|$.
Since there are $|V|$ vertices totally appearing $\sum_{v\in V}|T_{v}|$ times in $G$, each of vertices' occurrence has the same probability $p=\frac{1}{\sum_{v\in V}|T_{v}|}$ to be selected as the initial one.
This makes the sequences sampled by the following temporal random walk cover the entire graph evenly.

\begin{figure}
	\centerline{\includegraphics[width=0.3\textwidth]{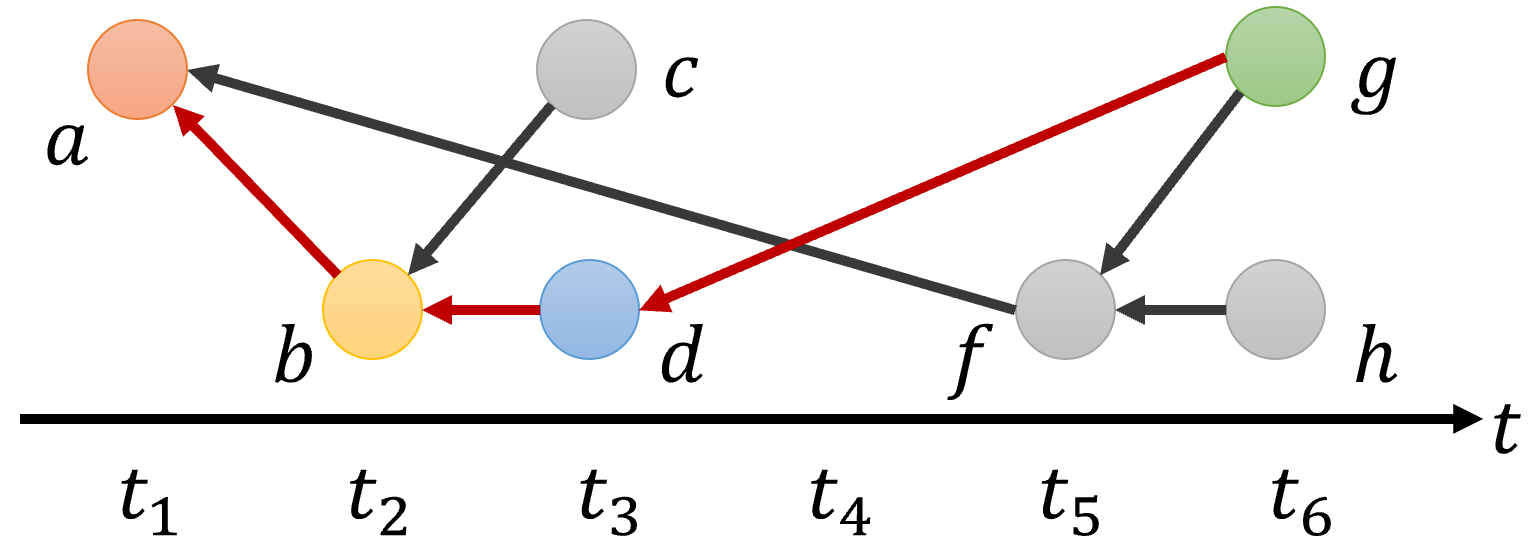}}
	\caption{An example of the time-centrality-biased temporal random walk starting from a vertex $a$. Since vertices $b$ and $f$ have the same degrees but ToE $\delta_{a,b}<\delta_{a,f}$, the walker will prefer $b$ more than $f$. Although vertices $c$ and $d$ have the same ToE, the degree of $d$ is higher than that of $c$, resulting in higher probability to be sampled by the walker after $b$.}
	\label{fig:rw_example}
\end{figure}

Next, we perform the temporal random walk to sample the connected vertices starting from every initial vertex.
The walker will only visit vertices whose ToV is greater than the previous one, making the sampled sequences be in line with the structural evolutions of $G$.
Usually, the upcoming vertices have higher chances to form edges with the high centrality ones \cite{gu2018rare}.
As the events/relationships happen in recent times may have stronger influence to the current vertex comparing to that of happened long time ago, we heavily sample vertices that are close in time to the current one.
In other words, the walker prefers to visit a vertex through the edge with a smaller ToE but a higher centrality.
Therefore, given a vertex $v^t$, our temporal random walk samples the next vertex $v_{next}^{t'}$ from its neighborhood set $\Gamma_{v^t}$ based on a transition probability by Eq. (\ref{Eq:Prob_RW}) following a time-centrality-biased distribution, where the ToV of every vertex in $\Gamma_{v^t}$ is greater than $t$, $degree(\cdot)$ is the vertex's degree for measuring its centrality, $\delta_{v,v^t}$ is the ToE of the edge connecting two vertices $v$ and $v^t$.
\begin{equation}
	\label{Eq:Prob_RW}
	p_{rw}(v_{next}^{t'})=\dfrac{degree(v_{next}^{t'})}{\sum_{v\in\Gamma_{v^t}}degree(v)}(1-\dfrac{\delta_{v_{next}^{t'},v^t}}{\sum_{v\in\Gamma_{v^t}}\delta_{v,v^t}})
\end{equation}
Fig.~\ref{fig:rw_example} shows a running example of the time-centrality-biased temporal random walk.
Leveraging the centrality of vertices and ToE to sample the dynamic graph makes the vertex distribution and its linkage evolutionary patterns in the generated corpus consistent with the original graph, thus providing comprehensive information for later embedding.

We record the first $\ell$ vertices that the walker visits to construct the edge sequences $s_t$, $t=1,2,\cdots,m$.
The ToV of the first vertex in $s_t$ exactly is the evolution starting time and the total sum of ToE of all edges in $s_t$ indicates the time the evolutions of $s_t$ lasts.
Finally, the dynamic graph $G$ has been modeled as a set of edge sequence $\hat{G}=\{s_1,s_2,\cdots,s_t,\cdots,s_m\}$.
Every edge sequence $s_t$ is a local structure consisting of $\ell$ vertices in the dynamic graph.
An example is shown on the bottom-right in Fig. \ref{fig:dynamic_graph_modeling} demonstrating that the time-centrality-biased temporal random walk is capable of capturing the asynchronous structural evolutions in the dynamic graph.

In order to easily facilitate the embedding algorithm, we further attach a virtual vertex $\left<EOS\right>$ with zero ToE to the end of every $s_t$.
If the walker early stops before reaching the maximum sampling number, we supplement $\left<EOS\right>$ at the end to make the sequence have the same length as others.

\section{Embedding Asynchronous Structural Evolutions in The Dynamic Graph}

In this section, we present the details of our proposed TADGE to embed the asynchronous structural evolutions from $\hat{G}=\{s_1,s_2,\cdots,s_t,\cdots,s_m\}$.
TADGE consists of three parts:
\textbf{(1) Edge Formation Embedding:} a Time-aware Transformer to embed the dynamic connection changes among vertices while preserving the ToE;
\textbf{(2) Structural Evolution Embedding:} a multi-head self-attention model to embed the asynchronous evolution starting time for every local structure in the dynamic graph;
and \textbf{(3) Representation Fusion:} encoding final vertex representation by fusing the above edge formation and structural evolution embeddings.

In sections~\ref{sec:embedding_dynamic_edge_ToE} to~\ref{sec:rep_fusion}, we first introduce how TADGE encodes the representation for a vertex $v_i$ in one of its occurring edge sequences $s_t\in\hat{G}$.
Then, in section~\ref{sec:training_TADGE}, we present the strategies for training the TADGE by using all edge sequences containing $v_i$ to update its representation.

In the rest of this paper, we use bold uppercase letters to denote matrices, bold lowercase letters to denote vectors, and non-bold characters to denote scalars if not clarified.

\subsection{Embedding Dynamic Edge Formation with ToE}
\label{sec:embedding_dynamic_edge_ToE}

When a vertex comes and forms a new edge, it actually is selecting an existing vertex in the graph to connect based on past connection evolution and ToE.
From the vertex point of view, the edge sequences carrying vertices' connection evolution can be regarded as the sequences of vertices linking by these edges.
Since the edges have different ToE, the time interval between consecutive vertices in the sequence varies from each other, thus bringing time-varying sequential dependency to vertices in the edge sequence.
Therefore, we embed the dynamic edge formation by learning (1) the time-varying sequential dependency among vertices, and (2) pairwise connections among vertices. 

\begin{figure}
	\centerline{\includegraphics[width=0.45\textwidth]{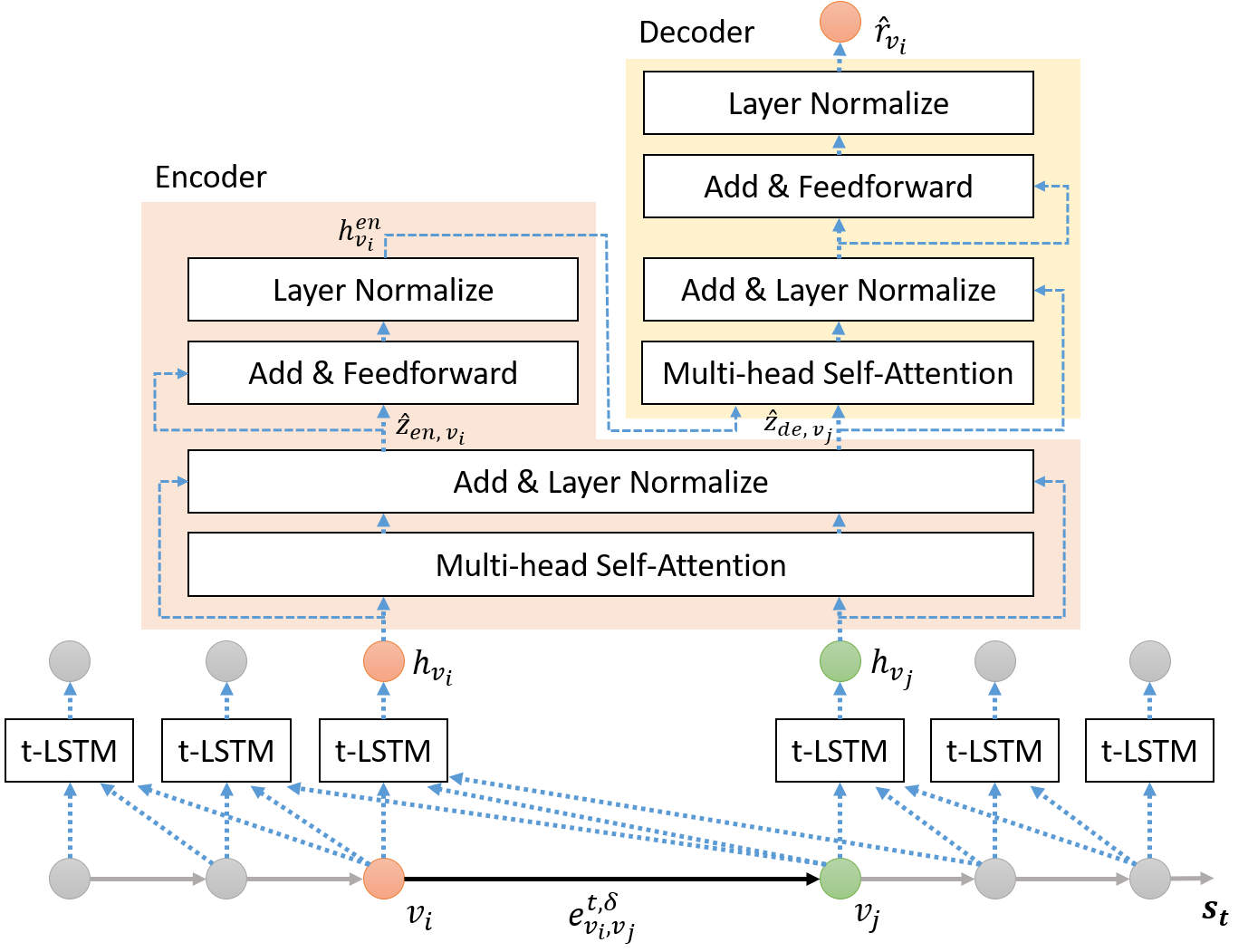}}
	\caption{An illustration of the Time-aware Transformer for embedding $v_i$.}
	\label{fig:ill_timeaware_transformer}
\end{figure}

Inspired by the R-Transformer \cite{wang2019r-Transformer}, we propose a Time-aware Transformer consisting of a t-LSTM model and a Transformer to respectively learn vertices' sequential dependency with ToE and their pairwise connections in every edge sequence $s_t$. The overall architecture is shown in Fig. \ref{fig:ill_timeaware_transformer} and $\boldsymbol{\hat{r}}_{v_i}$ is the obtained edge formation embedding for $v_i$.

\subsubsection{Learning Time-varying Sequential Dependency}
Let us denote the target vertex that we are going to embed as $v_i$.
To learn the sequential impact of ahead connected vertices and the ToE to $v_i$ in $s_t$, we build a t-LSTM model that starts from a standard LSTM unit and connects to time-aware LSTM units \cite{baytas2017patient}. 
It discounts the short-term effects from the long-term memory and supplements the impact of ToE between two consecutive nodes in the sequence, thus learning the time-varying sequential dependency.

Specifically, given the input information $\boldsymbol{x}_{v_i}\in \mathbb{R}^k$ of $v_i$ and its connected vertex $v_j$'s overall memory $\boldsymbol{c}_{v_j}$ and the hidden state $\boldsymbol{h}_{v_j}\in \mathbb{R}^k$, the time-aware LSTM unit first decomposes $\boldsymbol{c}_{v_j}$ into short-term memory $\boldsymbol{c}^S_{v_j}$ and long-term memory $\boldsymbol{c}^L_{v_j}$ as below:
\begin{align}
\boldsymbol{c}^S_{v_j} & = \tanh \left( \boldsymbol{W}_d \boldsymbol{c}_{v_j}+\boldsymbol{b}_d \right)\\
\boldsymbol{c}^L_{v_j} & = \boldsymbol{c}_{v_j}-\boldsymbol{c}^S_{v_j}
\end{align}
where $\{\boldsymbol{W}_d,\boldsymbol{b}_d\}$ are the trainable weights and bias for the subspace decomposition.
Next, we further decay the decayed short-term memory $\boldsymbol{c}^S_{v_j}$ by a heuristic function in Eq.~(\ref{Eq:decay_short_term_memory}) such that the longer ToE the fewer effects to the short-term memory. $\delta_{v_i,v_j}$ is the ToE of the edge linking $v_i$ and $v_j$ while $e$ is the Euler's number.
\begin{equation}
\label{Eq:decay_short_term_memory}
\boldsymbol{\hat{c}}^S_{v_j}=\frac{\boldsymbol{c}^S_{v_j}}{\log(e+\delta_{v_i,v_j})}
\end{equation}
Lastly, we compose the adjusted overall memory back by
\begin{equation}
\boldsymbol{\tilde{c}}_{v_j}=\boldsymbol{c}^L_{v_j} + \boldsymbol{\hat{c}}^S_{v_j}
\end{equation}
The rest parts in the time-aware LSTM unit are the same as the standard LSTM as shown below:
\begin{align}
\boldsymbol{f}_{v_i} &=\rm{sigmoid}(\boldsymbol{W}_f\boldsymbol{x}_{v_i}+\boldsymbol{U}_f\boldsymbol{h}_{v_j}+\boldsymbol{b}_f)\\
\boldsymbol{g}_{v_i} &=\rm{sigmoid}(\boldsymbol{W}_g\boldsymbol{x}_{v_i}+\boldsymbol{U}_g\boldsymbol{h}_{v_j}+\boldsymbol{b}_g)\\
\boldsymbol{o}_{v_i} &=\rm{sigmoid}(\boldsymbol{W}_o\boldsymbol{x}_{v_i}+\boldsymbol{U}_o\boldsymbol{h}_{v_j}+\boldsymbol{b}_o)\\
\boldsymbol{\bar{c}}_{v_i} &=\rm{sigmoid}(\boldsymbol{W}_c\boldsymbol{x}_{v_i}+\boldsymbol{U}_c\boldsymbol{h}_{v_j}+\boldsymbol{b}_c)\\
\boldsymbol{c}_{v_i} &=\boldsymbol{f}_{v_i}*\boldsymbol{\tilde{c}}_{v_j}+\boldsymbol{g}_{v_i}*\boldsymbol{\bar{c}}_{v_i}\\
\boldsymbol{h}_{v_i} &=\boldsymbol{o}_{v_i}*\tanh(\boldsymbol{c}_{v_i})
\end{align}
$\boldsymbol{c}_{v_i}$ is $v_i$'s current memory.
$\boldsymbol{h}_{v_i}$ is the output hidden state of $v_i$ carrying the time-varying sequential dependency between itself and vertices in $s_t$.
$*$ is the element-wise multiplication.
$\{\boldsymbol{W}_f,\boldsymbol{U}_f,\boldsymbol{b}_f\}$, $\{\boldsymbol{W}_g,\boldsymbol{U}_g,\boldsymbol{b}_g\}$, $\{\boldsymbol{W}_o,\boldsymbol{U}_o,\boldsymbol{b}_o\}$, and $\{\boldsymbol{W}_c,\boldsymbol{U}_c,\boldsymbol{b}_c\}$ are the trainable weights and bias of the forget gate $f$, input gate $g$, output gate $o$, and candidate memory $\bar{c}$. 
Consequently, our t-LSTM model is capable to learn the impact of past connections and ToE on the target vertex.

\subsubsection{Embedding Pairwise Connection between Vertices}
On top of the t-LSTM, we embed the direct connection between $v_i$ and $v_j$ by the self-attention mechanism in a Transformer architecture \cite{vaswani2017attention}. 
Intuitively, the edge formation process is to select an existing vertex from the graph for the upcoming one $v_i$ to link to.
In another word, $v_i$ is able to view the candidate vertices in the graph and then determines which one it will form the edge with.
We model this process by the self-attention mechanism that contains two steps.
First, we build an encoder to compute the self-attention between $v_i$ and vertices in the local structure $s_t$ that $v_i$ belongs.
It measures the impact of connected vertices in the local structure for $v_i$ to form edges.
Second, regarding the encoder output as the context information of forming the edges, we further build a decoder to learn which vertices $v_i$ exactly connects to, thus well embedding the dynamic edge formation.

\textbf{Encoder.}
Given the t-LSTM embeddings of $\ell$ vertices in the edge sequence $s_t$, denoting as $\boldsymbol{H}_v=[\boldsymbol{h}_{v_1},\boldsymbol{h}_{v_2},\cdots,\boldsymbol{h}_{v_{\ell}}]^T\in \mathbb{R}^{\ell\times k}$, we start with the multi-head self-attention module:
\begin{equation}
\label{Eq:encoder_atten}
\small
\boldsymbol{Z}^o_{en}={\rm softmax}\left(\frac{\boldsymbol{Q}^o_{en} {\boldsymbol{K}^o_{en}}^T}{\sqrt{k}}+\boldsymbol{M}\right)\boldsymbol{V}^o_{en}
\end{equation}
where $\boldsymbol{Z}^o_{en}\in \mathbb{R}^{\ell\times k}$ is the computed attention from the attention-head $o=1,2,\cdots,\bar{o}$ while $\sqrt{k}$ is a scaling factor to smooth the softmax calculation for avoiding extremely large values of the inner product.
$\boldsymbol{Q}^o_{en},\boldsymbol{K}^o_{en},\boldsymbol{V}^o_{en}\in \mathbb{R}^{\ell\times k}$ respectively represent the queries, keys, and values of the self-attention obtained by the linear projection of $\boldsymbol{H}_v$,
\begin{equation}
\small
\boldsymbol{Q}^o_{en}=\boldsymbol{H}_v \boldsymbol{W}^{Q,o}_{en}, \boldsymbol{K}^o_{en}=\boldsymbol{H}_v \boldsymbol{W}^{K,o}_{en},
\boldsymbol{V}^o_{en}=\boldsymbol{H}_v \boldsymbol{W}^{V,o}_{en},
\end{equation}
where $\boldsymbol{W}^{Q,o}_{en},\boldsymbol{W}^{K,o}_{en},\boldsymbol{W}^{V,o}_{en}\in \mathbb{R}^{k\times k}$ are trainable projection weights for the queries, keys, and values at every attention-head.
$\boldsymbol{M}\in \{0,-\infty\}^{\ell\times \ell}$ is a constant attention mask.
Since the vertex $v_i$ cannot see and form edges with $v_j$ that joins $s_t$ after $v_i$, we set $M_{i,j}=-\infty$ if $i<j$.
Therefore, the attention value between $v_i$ and not-yet-appeared vertex $v_j$ will be $0$ after softmax calculation in Eq. (\ref{Eq:encoder_atten}).
For the vertex $v_j$ that joins $s_t$ earlier than $v_i$, i.e., $i \geq j$, we disable the mask by setting $M_{i,j}=0$ and compute their attentions.
Unlike GAT~\cite{velickovic2018graph} which computes attention between directly connected vertices and merely preserves the first-order proximity of the graph structure, we learn the attention between the target vertex and other vertices in the local structure that the target one belongs.
Hence, our attention mechanism is capable of preserving both first-order and higher-order proximity of the graph structure.

Next, we concatenate all attentions obtained from every attention-head, denoting as $\boldsymbol{Z}_{en}$, and propagate low-layer queries to $\boldsymbol{Z}_{en}$ in residual connections with the layer normalization as shown in Eq. (\ref{Eq:encoder_atten_multihead}), which can improve the expressive capability and prevent from the vanishing gradient during training.
\begin{align}
\boldsymbol{Z}_{en} &= [\boldsymbol{Z}^1_{en}, \boldsymbol{Z}^2_{en},\cdots, \boldsymbol{Z}^{\bar{o}}_{en}]\in\mathbb{R}^{\ell\times \bar{o}k}\\
\boldsymbol{Q}_{en} &= [\boldsymbol{Q}^1_{en}, \boldsymbol{Q}^2_{en},\cdots, \boldsymbol{Q}^{\bar{o}}_{en}]\in\mathbb{R}^{\ell\times \bar{o}k}
\end{align}
\begin{equation}
\label{Eq:encoder_atten_multihead}
\boldsymbol{\hat{Z}}_{en}=LN(\boldsymbol{Z}_{en}+\boldsymbol{Q}_{en})=\boldsymbol{S}_{ln}*\frac{\boldsymbol{Z}_{en}+\boldsymbol{Q}_{en}-\mu}{\epsilon}+\boldsymbol{B}_{ln}
\end{equation}
$LN(\cdot)$ is the layer normalization function where $\mu$ and $\epsilon$ are the mean and variance of elements in the input tensor, and $\boldsymbol{S}_{ln}$ and $\boldsymbol{B}_{ln}$ are trainable scaling weights and bias for maintaining the representation ability of the network.
$*$ is the element-wise multiplication.

Lastly, we fuse the multi-head attentions $\boldsymbol{\hat{Z}}_{en}$ by a two-layer fully connected feed-forward network (FFN) with a non-linear activation ${\rm ReLU}(\cdot)=\max\{0,\cdot\}$ under the residual connection setting and following by the layer normalization to obtain the attention embedding $\boldsymbol{H}^{en}_v$:
\begin{equation}
\label{Eq:encoder_atten_embedding}
\boldsymbol{H}^{en}_v=LN({\rm ReLU}(\boldsymbol{\hat{Z}}_{en}\boldsymbol{W}^1_{f}+\boldsymbol{B}^1_{f}+\boldsymbol{\hat{Z}}_{en})\boldsymbol{W}^2_{f}+\boldsymbol{B}^2_{f})
\end{equation}
where $\boldsymbol{W}^1_{f} \in\mathbb{R}^{\bar{o}k\times \bar{o}k}$, $\boldsymbol{W}^2_{f} \in\mathbb{R}^{\bar{o}k\times k}$,  $\boldsymbol{B}^1_{f} \in\mathbb{R}^{\ell \times \bar{o}k}$ and $\boldsymbol{B}^2_{f} \in\mathbb{R}^{\ell \times k}$ are the trainable weights and bias of the FFN.
Each row in $\boldsymbol{H}^{en}_v$ corresponds to the encoding of a vertex in $s_t$.

\textbf{Decoder.}
The goal of a decoder is to learn which vertices the encoding one connects to so that an edge forms.
Thus, the input of decoder is the t-LSTM embeddings of the second vertex to the last one in $s_t$, denoting as $\boldsymbol{H}'_v=[\boldsymbol{h}_{v_2},\boldsymbol{h}_{v_3},\cdots,\boldsymbol{h}_{v_{\ell}},\boldsymbol{h}_{EOS}]^T\in \mathbb{R}^{\ell\times k}$, and their attentions $\boldsymbol{\hat{Z}}'_{en}$ obtained by the encoder through Eq. (\ref{Eq:encoder_atten}) to (\ref{Eq:encoder_atten_multihead}).

To further learn the pair-wise connection in $s_t$, we build another self-attention module, employing $\boldsymbol{\hat{Z}}'_{en}$ as queries, $\boldsymbol{H}^{en}_v$, which is the self-attentive embeddings of source vertices, as keys $K^o_{de}$ and values $V^o_{de}$ as showed in Eq. (\ref{Eq:decoder_atten}),
\begin{equation}
\small
\label{Eq:decoder_atten}
\boldsymbol{Q}^o_{de}=\boldsymbol{\hat{Z}}'_{en} \boldsymbol{W}^{Q,o}_{de}, \boldsymbol{K}^o_{de}=\boldsymbol{H}^{en}_v \boldsymbol{W}^{K,o}_{de} ,\boldsymbol{V}^o_{de}=\boldsymbol{H}^{en}_v \boldsymbol{W}^{V,o}_{de}
\end{equation}
where $\boldsymbol{W}^{Q,o}_{de}\in \mathbb{R}^{\bar{o}k\times k}$ and $\boldsymbol{W}^{K,o}_{de},\boldsymbol{W}^{V,o}_{de}\in \mathbb{R}^{k\times k}$ are trainable projection weight matrices at every attention-head $o=1,2,\cdots,\bar{o}$. 
Then, we calculate their self-attention by:
\begin{align}
\boldsymbol{Z}^o_{de} &= {\rm softmax}\left(\frac{\boldsymbol{Q}^o_{de} {\boldsymbol{K}^o_{de}}^T}{\sqrt{k}}+\boldsymbol{M}\right)\boldsymbol{V}^o_{de}\\
\boldsymbol{Z}_{de} &= [\boldsymbol{Z}^1_{de}, \boldsymbol{Z}^2_{de},\cdots, \boldsymbol{Z}^{\bar{o}}_{de}]\in\mathbb{R}^{\ell\times \bar{o}k}\\
\boldsymbol{Q}_{de} &= [\boldsymbol{Q}^1_{de}, \boldsymbol{Q}^2_{de},\cdots, \boldsymbol{Q}^{\bar{o}}_{de}]\in\mathbb{R}^{\ell\times \bar{o}k}\\
\boldsymbol{\hat{Z}}_{de} &= LN(\boldsymbol{Z}_{de}+\boldsymbol{Q}_{de})
\end{align}
Lastly, a two-layer fully connected FFN with ReLU activation is built to obtain the edge formation embedding $\boldsymbol{\hat{R}}_v$:
\begin{equation}
\boldsymbol{\hat{R}}_v=LN({\rm ReLU}(\boldsymbol{\hat{Z}}_{de}\boldsymbol{W}^3_{f}+\boldsymbol{B}^3_{f}+\boldsymbol{\hat{Z}}_{de})\boldsymbol{W}^4_{f}+\boldsymbol{B}^4_{f})
\end{equation}
where $\boldsymbol{W}^3_{f} \in\mathbb{R}^{\bar{o}k\times \bar{o}k}$, $\boldsymbol{W}^4_{f} \in\mathbb{R}^{\bar{o}k\times k}$,  $\boldsymbol{B}^3_{f} \in\mathbb{R}^{\ell \times \bar{o}k}$ and $\boldsymbol{B}^4_{f} \in\mathbb{R}^{\ell \times k}$ are trainable weights and bias of this FFN.
Each row in $\boldsymbol{\hat{R}}_v$ corresponds to the edge formation embedding of a vertex in $s_t$.

Since the vertices' sequential dependency and ToE have been well preserved by the t-LSTM module and the encoder-decoder merely needs to learn the pairwise connection between vertices in $s_t$, it is not necessary to incorporate with the position embedding which is used in the original Transformer.
In order to further boost the representation power of our Time-aware Transformer model, we respectively stack $N$ blocks of the multi-head attention in both encoder and decoder.
We adopt dropout on every residual FFN as a regularization strategy to prevent the model from over-fitting.
Eventually, our Time-aware Transformer model is capable of simultaneously embedding the edges connecting pairs of vertices together with their ToE.

\subsection{Structure Embedding with Evolution Starting Time}
\label{sec:embed_struc_evolving}

When we treat every edge sequence $s_t$ as a whole, the dynamic graph becomes a sequence $\hat{G}=\{s_1,\cdots,s_t,\cdots,s_m\}$ representing the evolving time series of the local structures.
This new sequence carries the global evolution patterns of $G$ that consists of the asynchronous evolving of every local structure.
We are going to learn a representation vector $\boldsymbol{r}_{s_t}$ for the local structure $s_t$, where $t=1,2,\cdots,m$.
The structure embedding $\boldsymbol{r}_{s_t}$ should preserve the sequential dependency of every local structure in $\hat{G}$ and its evolution starting time which exactly is the ToV of its first vertex.

Let's denote the edge formation embedding of a vertex $v_i$ as $\boldsymbol{\hat{r}}_{v_i}$ that corresponds to a row vector in $\boldsymbol{\hat{R}}_v$.
We first aggregate $\boldsymbol{\hat{r}}_{v_i}$ for every $v_i$ in $s_t$ to get the initial structure embedding $\boldsymbol{h}_{s_t}$ as shown in Eq. (\ref{Eq:initial_struc_embedding}).
Since the virtual vertex $\left<EOS\right>$ appears at the end of every $s_t$ and does not contribute anything to the structures, we merely count the first $\ell$ vertices in $s_t$ to get $\boldsymbol{h}_{s_t}$ regardless they are $\left<EOS\right>$ or not.
For $\forall s_t\in \hat{G}$, the initial structure embeddings can be written into a matrix form $\boldsymbol{H}_s=[\boldsymbol{h}_{s_1},\boldsymbol{h}_{s_2},\cdots,\boldsymbol{h}_{s_m}]^T \in\mathbb{R}^{m\times k}$.
\begin{equation}
\label{Eq:initial_struc_embedding}
\boldsymbol{h}_{s_t}=\sum_{v_i\in s_t}\boldsymbol{\hat{r}}_{v_i}.
\end{equation}

We again employ the self-attention to learn the sequential relationship among local structures as shown below:
\begin{equation}
\boldsymbol{Q}^o_{s}=\boldsymbol{H}_s \boldsymbol{W}^{Q,o}_{s}, \boldsymbol{K}^o_{s}=\boldsymbol{H}_s \boldsymbol{W}^{K,o}_{s},
\boldsymbol{V}^o_{s}=\boldsymbol{H}_s \boldsymbol{W}^{V,o}_{s}
\end{equation}
\begin{align}
\boldsymbol{Z}^o_{s} &= {\rm softmax}\left(\frac{\boldsymbol{Q}^o_{s} {\boldsymbol{K}^o_{s}}^T}{\sqrt{k}}\right)\boldsymbol{V}^o_{s}\\
\boldsymbol{Z}_{s} &= [\boldsymbol{Z}^1_{s}, \boldsymbol{Z}^2_{s},\cdots, \boldsymbol{Z}^{\bar{o}}_{s}]\in\mathbb{R}^{m\times \bar{o}k}\\
\boldsymbol{Q}_{s} &= [\boldsymbol{Q}^1_{s}, \boldsymbol{Q}^2_{s},\cdots, \boldsymbol{Q}^{\bar{o}}_{s}]\in\mathbb{R}^{m\times \bar{o}k}\\
\boldsymbol{\hat{Z}}_{s} &= LN(\boldsymbol{Z}_{s}+\boldsymbol{Q}_{s})
\end{align}
where $\boldsymbol{Q}^o_{s},\boldsymbol{K}^o_{s},\boldsymbol{V}^o_{s}\in \mathbb{R}^{m\times k}$ respectively represent the queries, keys, and values obtained by the linear projection of $\boldsymbol{H}_s$.
$\boldsymbol{W}^{Q,o}_{s},\boldsymbol{W}^{K,o}_{s},\boldsymbol{W}^{V,o}_{s}\in \mathbb{R}^{k\times k}$ are trainable projection weight matrices at every attention-head $o=1,2,\cdots,\bar{o}$.
We again build a two-layer fully connected FFN with the ReLU activation to obtain the structure embedding $\boldsymbol{\hat{R}}_s$:
\begin{equation}
\boldsymbol{\hat{R}}_s=LN({\rm ReLU}(\boldsymbol{\hat{Z}}_{s}\boldsymbol{W}^5_{f}+\boldsymbol{B}^5_{f}+\boldsymbol{\hat{Z}}_{s})\boldsymbol{W}^6_{f}+\boldsymbol{B}^6_{f})
\end{equation}
where $\boldsymbol{W}^5_{f} \in\mathbb{R}^{\bar{o}k\times \bar{o}k}$, $\boldsymbol{W}^6_{f} \in\mathbb{R}^{\bar{o}k\times k}$, $\boldsymbol{B}^5_{f} \in\mathbb{R}^{m \times \bar{o}k}$ and $\boldsymbol{B}^6_{f} \in\mathbb{R}^{m \times k}$ are the trainable weights and bias of the FFN.

To this end, we obtain the structure embedding $\boldsymbol{\hat{r}}_{s_t}$ for every $s_t$ in $\hat{G}$, which is corresponding to the $t$th row of $\boldsymbol{\hat{R}}_s$, preserving the sequential relationship among every local structure in $\hat{G}$.
Different from embedding the edge formation, we do not build a decoder here because local structures in $\hat{G}$ do not have strict pair-wise relationships and the encoder has already been able to learn their sequential relationships pretty well.

To encode the asynchronous evolution timing of every local structure into the structure embedding $\boldsymbol{\hat{R}}_s$, we train the above attention model by a regression task, estimating the evolution time interval $\delta_{s_t,s_{t'}}$ between any two local structures $s_t$ and $s_{t'}$.
To keep the article organization consistent, we present the details of the regression task and discuss the criteria of improving the training efficiency in Section \ref{sec:training_task_evo_time_interval}.

\subsection{Representation Fusion}
\label{sec:rep_fusion}

To obtain the representation of vertex $v_i$ in one of its occurring edge sequences $s_t$, we fuse $v_i$'s edge formation embedding $\boldsymbol{\hat{r}}_{v_i}$ together with its structure embedding $\boldsymbol{\hat{r}}_{s_t}$ by summing them up and get $\boldsymbol{\acute{r}}_{v_i}=\boldsymbol{\hat{r}}_{v_i}+\boldsymbol{\hat{r}}_{s_t}$.
Then, we input it into a FFN with the $ReLU(\cdot)$ activation to encode the final vertex representation as shown in Eq. (\ref{Eq:global_local_fusion}).
\begin{equation}
\label{Eq:global_local_fusion}
\boldsymbol{r_{v_i}}=LN({\rm ReLU}(\boldsymbol{\acute{r}}_{v_i}^T\boldsymbol{W}^7_{f}+\boldsymbol{b}^7_{v_i}+\boldsymbol{\acute{r}}_{v_i}^T)\boldsymbol{W}^8_{f}+\boldsymbol{b}^8_{v_i})
\end{equation}
$\boldsymbol{W}^7_{f}, \boldsymbol{W}^8_{f} \in\mathbb{R}^{k\times k}$ and $\boldsymbol{b}^7_{v_i}, \boldsymbol{b}^8_{v_i} \in\mathbb{R}^{1 \times k}$ are the trainable weights and bias of this FFN.
Since each vertex $v_i$ will likely appear in multiple edge sequences, we will present how to train the TADGE by using all edge sequences containing $v_i$ to update its representation in next subsection.

\subsection{Training TADGE}
\label{sec:training_TADGE}

In order to effectively embed the asynchronous structural evolution, we train the TADGE model by three self-supervised tasks jointly.

\subsubsection{Training Task 1: Self-identification for Edge Formation Embedding}
\label{sec:training_task_self-identification}

We employ a vertex classification task to train the Time-aware Transformer for learning the edge formation embedding.
Intuitively, no matter how $v_i$'s connections change, its embedding $\boldsymbol{\hat{r}}_{v_i}$ should well identify $v_i$ itself since it is representing $v_i$'s edge formation information instead of other vertices.
We employ a softmax with cross-entropy loss to train the Time-aware Transformer well-classifying vertices' own identity as shown in Eq. (\ref{Eq:self_classification_loss}).
\begin{equation}
\small
\label{Eq:self_classification_loss}
\begin{gathered}
\mathcal{L}_{v} = -\sum_{v_i\in s_t}\sum_{s_t\in \hat{G}}\left(\boldsymbol{y}_{v_i}^T\log\left({\rm softmax}\left(\boldsymbol{\hat{R}}_v\boldsymbol{\hat{r}}_{v_i}\right)\right)\right.\\
\left.+\left(1-\boldsymbol{y}_{v_i}^T\right)\log\left(1-{\rm softmax}\left(\boldsymbol{\hat{R}}_v\boldsymbol{\hat{r}}_{v_i}\right)\right)\right)
\end{gathered}
\end{equation}
$\boldsymbol{\hat{R}}_v$ contains the edge formation embeddings of all vertices in $s_t$.
$\boldsymbol{y}_{v_i}\in\mathbb{R}^{|V|\times 1}$ is the self-identification of $v_i$ in one-hot encoding that the $i$th element in $\boldsymbol{y}_{v_i}$ is $1$ and others are $0$.

\subsubsection{Training Task 2: Evolution Time Interval Regression for Structural Evolution Embedding}
\label{sec:training_task_evo_time_interval}

Embedding the evolution starting time is the key to preserve the asynchronous structural evolution.
We employ a regression task on approximating the evolution time interval $\delta_{s_t,s_{t'}}$ between any pair of local structures $s_t$ and $s_{t'}$ while learning their structure embeddings $\boldsymbol{\hat{r}}_{s_t}$ and $\boldsymbol{\hat{r}}_{s_{t'}}$ in $\boldsymbol{\hat{R}}_s$.
Mathematically, this regression task is written in Eq. (\ref{eq:task_struc_time_interval_reg}), where $\boldsymbol{w}_s\in\mathbb{R}^{k\times 1}$ is the trainable linear projection weights.

\begin{equation}
\small
\label{eq:task_struc_time_interval_reg}
\begin{gathered}
\mathcal{L}_{s} = \frac{1}{m(m-1)}\sum_{s_t\in \hat{G}}\sum_{\begin{subarray}{c}s_{t'}\in \hat{G}\\ t'<t\end{subarray}} \left(\boldsymbol{w}_s^T\left(\boldsymbol{\hat{r}}_{s_t}+\boldsymbol{\hat{r}}_{s_{t'}}\right) - \delta_{s_t,s_{t'}}\right)^2
\end{gathered}
\end{equation}

The benefits of regressing the evolution time interval are twofold.
First, it well preserves the sequential dependency among local structures while merely embedding the absolute evolution starting time cannot achieve.
Second, it augments the scale of training samples so that the model is easy to converge and avoids under-fitting.
However, there are $m(m-1)/2$ pairs of local structures being used to compute the gradient of Eq. (\ref{eq:task_struc_time_interval_reg}).
When $m$ becomes extremely large, the gradient calculation will take a long time, which is an efficiency bottleneck of training the TADGE.

Inspired by the negative sampling, we adopt a sliding window with length $\ell_s$ to sample the structure sequence $\hat{G}=\{s_1,s_2,\cdots,s_m\}$ and construct structure pairs within the siding window as training samples.
The sliding step size $\tilde{o}$ should satisfy $1\leq \tilde{o} < \ell_s$ for ensuring two consecutive subsequences having overlap.
Otherwise, the continuous evolution of the whole dynamic graph will be intermittent.
Thus, the maximum number of sliding windows $m_w$ satisfies $m_w\leq m-\ell_s+1$.
Since there are $\ell_s(\ell_s-1)/2$ training samples in each sliding window, $m_w\ell_s(\ell_s-1)/2$ training samples are obtained in total.
According to the Lemma \ref{Lemma:num_global_sampling} and \ref{Lemma:upper_bround_global_sampling}, when the sliding windows satisfy $\ell_s^2-3\ell_s+3<m$, the scale of training samples will definitely be reduced while ensuring the overlap of sliding windows, therefore improving the training efficiency.

\begin{lemma}
	\label{Lemma:num_global_sampling}
	$m_w\frac{\ell_s(\ell_s-1)}{2} < \frac{m(m-1)}{2}$ when $m_w<\frac{(m-1)^2}{(\ell_s-1)^2}$.
\end{lemma}

\begin{proof}
	Since $m> \ell_s > 1$, we have $\frac{m}{\ell_s}<\frac{m-1}{\ell_s-1}$. From $m_w\frac{\ell_s(\ell_s-1)}{2} < \frac{m(m-1)}{2}$, we get $m_w< \frac{m(m-1)}{\ell_s(\ell_s-1)}<\frac{(m-1)^2}{(\ell_s-1)^2}$.
\end{proof}

\begin{lemma}
	\label{Lemma:upper_bround_global_sampling}
	When $m>\ell_s>1$, $m_w\leq m-\ell_s+1$ and $m_w\frac{\ell_s(\ell_s-1)}{2} < \frac{m(m-1)}{2}$ simultaneously establish if $\ell_s^2-3\ell_s+3<m$.
\end{lemma}

\begin{proof}
	Since $m-\ell_s+1$ is the upper bound of $m_w$ and $m_w\frac{\ell_s(\ell_s-1)}{2} < \frac{m(m-1)}{2}$ established when $m_w<\frac{(m-1)^2}{(\ell_s-1)^2}$, we let $m-\ell_s+1 < \frac{(m-1)^2}{(\ell_s-1)^2}$ and get $m-\ell_s<\frac{(m-1)^2}{(\ell_s-1)^2} - 1 = \frac{(m-\ell_s)(m+\ell_s-2)}{(\ell_s-1)^2}$. Therefore, $(\ell_s-1)^2< m+\ell_s-2$, and, finally, we have $\ell_s^2-3\ell_s+3<m$.
\end{proof}

\subsubsection{Training Task 3: Time-aware Edge Reconstruction for Final Representations}
\label{sec:training_task_time-aware_edge}

Edge reconstruction task has been widely adopted to train the static graph embedding algorithms.
It assumes that the representation of connected vertices should be close.
However, this assumption oversimplified the edge formation process in the dynamic graph since the temporal information such as ToV and ToE has been neglected.
A vertex can join the dynamic graph many times forming edges with the same pair of vertices but having different ToV and ToE.
In the dynamic graph, when vertices are connected and have similar ToV and ToE, their representation should be close.

We propose a new task, namely time-aware edge reconstruction, to simultaneously estimate the ToE while reconstructing the edges between pairs of vertices.
Given the final representation $\boldsymbol{R}_v=[\boldsymbol{r}_{v_1},\boldsymbol{r}_{v_2},\cdots]^T\in\mathbb{R}^{|V|\times k}$, the time-aware edge reconstruction not only classifies whether there is an edge between any pair of vertices, but also regresses the corresponding ToE at the same time.
The objective function is shown in Eq. (\ref{eq:task_Taware_edgRec}), where $\mathcal{L}_{edg}$ and $\mathcal{L}_{ToE}$ are for the edge reconstruction and the ToE regression, respectively.
\begin{equation}
\label{eq:task_Taware_edgRec}
\mathcal{L}_{r}=\mathcal{L}_{edg}+\mathcal{L}_{ToE}
\end{equation}
Similar to identifying the vertex itself in Section \ref{sec:training_task_self-identification}, we again build a softmax with cross-entropy loss for edge reconstruction as shown in Eq. (\ref{eq:task_Taware_edgRec_edgClf}), where $y_{v_i,v_j}=1$ if there is an edge between $v_i$ and $v_j$ in $s_t$, otherwise $y_{v_i,v_j}=0$.
\begin{equation}
	\small
	\label{eq:task_Taware_edgRec_edgClf}
	\begin{gathered}
			\mathcal{L}_{edg} = -\sum_{v_i,v_j\in s_t}\sum_{s_t\in \hat{G}}\left(y_{v_i,v_j}\log\left({\rm softmax}\left(\boldsymbol{r}_{v_i}^T \boldsymbol{r}_{v_j}\right)\right)\right.\\
		\left.+\left(1-y_{v_i,v_j}\right)\log\left(1-{\rm softmax}\left(\boldsymbol{r}_{v_i}^T \boldsymbol{r}_{v_j}\right)\right)\right)
	\end{gathered}
\end{equation}
\begin{equation}
\label{eq:task_ToE_reg}
\small
\mathcal{L}_{ToE} = \frac{1}{2\hat{m}}\sum_{v_i,v_j\in s_t}\sum_{s_t\in \hat{G}} \left(\boldsymbol{w}_{ToE}^T\left(\boldsymbol{r}_{v_i}+\boldsymbol{r}_{v_j}\right) - \delta_{v_i,v_j}\right)^2
\end{equation}
$\mathcal{L}_{ToE}$ in Eq. (\ref{eq:task_ToE_reg}) is a ToE regression that approximates the ToE $\delta_{v_i,v_j}$ of the edge linking $v_i$ and $v_j$ by using their representations $\boldsymbol{r}_{v_i}$ and $\boldsymbol{r}_{v_j}$.
$\boldsymbol{w}_{ToE}\in\mathbb{R}^{k\times 1}$ is the trainable linear projection weights, and $\hat{m}$ is the number of training edges.
When we minimize $\mathcal{L}_{r}$ in Eq. (\ref{eq:task_Taware_edgRec}), the optimal results will be obtained if and only if both the edge reconstruction loss in Eq. (\ref{eq:task_Taware_edgRec_edgClf}) and the ToE regression loss in Eq. (\ref{eq:task_ToE_reg}) reach the minimum, eventually making vertices' representation close if they are connected and have similar ToV and ToE.

\subsubsection{Optimization Strategy}
As TADGE is built upon the deep neural network structure, we initialize the representation $R_v$ by using DeepWalk \cite{perozzi2014deepwalk} and then apply Adam, a mini-batch stochastic gradient descent optimizer, to learn the model parameters by minimizing the joint loss:
\begin{equation}
	\label{Eq:loss}
	\mathcal{L}=\mathcal{L}_{v}+\mathcal{L}_{s}+\mathcal{L}_{r}
\end{equation}
The DeepWalk is trained by a static graph constructed from the edges in the training set.
Thanks to the above three training tasks, the asynchronous structural evolution will be gradually embedded into the vertex representation.
Meanwhile, we normalize the ToE and the time interval between structures to $[0,1)$ by an arc-cotangent function $\delta =2\arctan(\delta)/\pi$, suppressing the effects of a very large value of $\delta$ on the model convergence.

It is worth mentioning that adding weighting factors to each sub-objective in Eq. (\ref{Eq:loss}) can balance the scale difference of their effects on the overall loss and result in better performance.
However, the weighting factors are application-specific.
Therefore, in this work, we assign equal weights to each sub-objective, training a general model regardless of the specific applications.
In addition, negative sampling can be applied to the sub-objectives $\mathcal{L}_{v}$ in Eq. (\ref{Eq:self_classification_loss}) and $\mathcal{L}_{edg}$ in Eq. (\ref{eq:task_Taware_edgRec_edgClf}) to improve the training efficiency.

\subsubsection{Training Protocol}
To properly train the TADGE, we feed all edge sequences $\{s_1,s_2,\cdots,s_t,\cdots,s_m\}$ into the TADGE one by one to update its trainable parameters.
Every time when an edge sequence $s_t$ is fed into the TADGE, we update the representation of vertices in the $s_t$ while keeping that of vertices not in the $s_t$ unchanged.
As a result, when the TADGE is trained on all edge sequences, each vertex $v_i$'s representation will be updated by every edge sequence where $v_i$ is in.

When updating the representation of a vertex $v_i$ in an edge sequence $s_t$, we must ensure that $v_i$ can merely see those vertices that joined the $s_t$ before $v_i$.
Therefore, we introduce the attention mask $\boldsymbol{M}$ in Eq. (\ref{Eq:encoder_atten}) and (\ref{Eq:decoder_atten}) when learning the edge formation embedding.
This requirement must also be met when learning the structure embedding for $v_i$.
To achieve this, we set $\boldsymbol{\hat{r}}_{v_j}=\boldsymbol{\hat{r}}_{v_{EOS}}$ for every $v_j\in s_t$ if $i<j\leq \ell$ when computing $\boldsymbol{h}_{s_t}$ in Eq. (\ref{Eq:initial_struc_embedding}) before learning the structure embedding for $v_i$, thus ensuring only existed local structures are visible to $v_i$.

\section{Experiments}

In this section, we validate the effectiveness of our proposed TADGE in three public real-world datasets and benchmark against the state-of-the-art baseline methods on several data mining applications.

\subsection{Experimental Setting}

\subsubsection{Datasets}
We benchmark the TADGE algorithm in three public real-world datasets, whose properties are introduced below.

\begin{itemize}
	\item \textbf{Transaction}\footnote{\url{https://snap.stanford.edu/data/soc-sign-bitcoin-otc.html}}\cite{kumar2016edge}.
		This is a dynamic bitcoin transaction network on the Bitcoin OTC platform.
		A vertex is a trader who buys and sells bitcoins.
		Two traders form an edge when they complete a transaction.
		The ToE is the time interval between buying and selling.
		Each trader is associated with a trustworthy label in low, middle, and high for classification.
	\item \textbf{Hyperlink}\footnote{\url{https://snap.stanford.edu/data/soc-RedditHyperlinks.html}}\cite{kumar2018community}.
		This is a dynamic subreddit-to-subreddit hyperlink network extracted from the posts that create hyperlinks from one subreddit to another on Reddit.
		The vertex is a subreddit and the edge is a hyperlink connecting two subreddits.
		Each vertex has a binary semantic label for classification.
	\item \textbf{Discussion}\footnote{\url{https://snap.stanford.edu/data/sx-superuser.html}}\cite{paranjape2017motifs}.
		It is a dynamic discussion network extracted from a stack exchange website.
		The vertex is a user who posts, replies, comments and answers questions on the website.
		Once a user interact with others, an edge is formed between them.
\end{itemize}
The statistics of these datasets are presented in Table \ref{tab:dataset}.
\begin{table}[htbp]
\caption{Statistics of datasets}
\centering
\resizebox{0.48\textwidth}{!}{
\label{tab:dataset}
\begin{tabular}{ccccccc}
\toprule
Dataset & $|\textbf{V}|$ & $|\textbf{E}|$ & Mean Degree & Mean ToE & Std. ToE & \#Classes\\
\midrule
Transaction & 5,881 & 35,592 & 3.665 & 30.693 days & 72.848 & 3 \\
Hyperlink & 54,075 & 571,927 & 7.701 & 58.350 days & 121.937 & 2 \\
Discussion & 194,085 & 1,443,339 & 3.987 & 76.575 days & 218.925 & - \\
\bottomrule
\end{tabular}}
\end{table}

\subsubsection{Baseline Methods}
We first choose the popular random-walk-based graph embedding methods and graph neural network models as the common baselines.
Both of them can only embed synchronous structural evolutions.
Besides, we replace the Time-aware Transformer and the structure embedding in our TADGE with the graph attention network (GAT) \cite{velickovic2018graph} to compare the effectiveness of our TADGE against GAT while embedding the asynchronous structural evolutions.
Lastly, we further evaluate the vertex representation merely learned by the Time-aware Transformer for exploring how TADGE benefits from the structure embedding.
\begin{itemize}
	\item \textbf{DeepWalk}\footnote{https://github.com/thunlp/OpenNE}\cite{perozzi2014deepwalk}.
	DeepWalk is a classical static graph embedding approach based on the random-walk and a skip-gram algorithm.
	\item \textbf{CTDNE} \cite{nguyen2018continuous}.
	The Continuous-Time Dynamic Network Embedding consists of a temporal random walk and a skip-gram algorithm to embed the continuous-time dynamic graph.
	It merely embeds the synchronous sequential edge formation but neglects the evolution timing and duration. 
	\item \textbf{GraphSAGE}\footnote{https://github.com/williamleif/GraphSAGE}\cite{hamilton2017graphSAGE}.
	This is a graph neural network approach for embedding the dynamic graph by computing the graph convolution from the vertices' connection change over time.
	We test different aggregators including GCN, mean, mean-pooling, and LSTM and report the best results in each dataset.
	\item \textbf{GAT}\footnote{https://github.com/PetarV-/GAT} \cite{velickovic2018graph}.
	This is one of the most popular attention-based approach for graph embedding, expanding the perception range of vertices from their local neighborhoods to all vertices in the whole graph.
	In order to benchmark our TADGE against GAT, We train the model with a loss function $\mathcal{L}=\mathcal{L}_{v}+\mathcal{L}_{r}$ so that the dynamic edge formation and the ToE are preserved by the GAT for comparison.
	\item \textbf{GAT-strc}
	We supplement the structure embedding presented in Section \ref{sec:embed_struc_evolving} into the above GAT, thus making it preserve the asynchronous structural evolutions.
	We train the model by using the same loss function $\mathcal{L}$ as the TADGE for a fair comparison.	
	\item \textbf{TADGE-tTran}
	We learn the vertex representation by the Time-aware Transformer which merely preserves the dynamic edge formation with the ToE.
	We remove the structural attention from the TADGE while setting $\boldsymbol{\hat{r}}_{s_t}=0$ and $\mathcal{L}_{s}=0$ to train the model.	
\end{itemize}


We do not benchmark the TADGE against those methods regarding the SGS as inputs.
Our graph model contains finer-grained edge information than the snapshot graphs.
Due to the time granularity issues of the SGS, we fail to construct the same training and test sets as TADGE for a fair comparison.

\subsubsection{Experiment Setup}

To split training and test sets, we first randomly select a set of vertices appearing at $m$ different timestamps following the uniform distribution. 
Next, we randomly select $20\%$ of them as the starting vertices to sample the dynamic graph $G$ by using the time-centrality-biased temporal random walk presented in Section \ref{sec:temporal_walk}.
The obtained edge sequences will be treated as the test set.
After that, we remove the test edges from the dynamic graph and sample it again starting from the rest $80\%$ of vertices to construct a training set, eventually ensuring that no test edges occur in the training set and all edge sequences are different in both training and test sets.
Due to different scales of datasets, we configure the temporal random walk with the settings in Table \ref{tab:parameter_temporal_walk}.

\begin{table}[h]
	\caption{Parameter Setting of Temporal Random Walk}
	\centering
	\label{tab:parameter_temporal_walk}
	\begin{tabular}{cccc}
		\toprule
		Dataset & $m$ & Max. $\ell$ & Min. $\ell$\\
		\midrule
		Transaction & 10,000 & 5 & 3 \\
		Hyperlink & 200,000 & 5 & 3 \\
		Discussion & 300,000 & 10 & 3 \\
		\bottomrule
	\end{tabular}
\end{table}

We set the minimum walk length to 3 so that there are at least 2 edges in every edge sequence indicating the structural evolutions.
Besides, we found that vertices have limited multi-hop connections in the dynamic graph, which is different from that of in a static graph.
Over $98\%$ of multi-hop connections in all three datasets do not exceed the maximum walk length.
We construct static graphs from the edges respectively in the training and test sets to train and test the DeepWalk and GraphSAGE.
Since the first step in CTDNE and GAT is sampling the dynamic graph as edge sequences, we skip this step and directly train them with the edge sequences in our obtained training set.
Hence, TADGE and all baselines are trained by the same set of edges or edge sequences for a fair comparison.

All experiments are conducted with $k=128$ as the dimension of representation vectors for the TADGE and all baseline methods in three datasets.
Following the recommended parameter setting of the Transformer in \cite{vaswani2017attention}, we set the number of head $\bar{o}=8$ and stack $N=6$ blocks for the Time-aware Transformer and GAT to learn the embeddings in the Hyperlink and Discussion datasets.
Due to the small scale of the Transaction data, we set $\bar{o}=4$ and $N=3$.
In training, each batch contains $200$, $500$, and $300$ edge sequences for Transaction, Hyperlink, and Discussion, respectively.
The learning rate of the Adam gradient descent optimizer is set to $0.005$.
We adopt the optimal parameters for DeepWalk, CTDNE and GraphSAGE, which presented in the original papers. 
Both TADGE and baseline methods are trained with enough epochs for ensuring the convergence. 
All experiments are conducted on a standard workstation with Intel Xeon Gold 5122 CPUs, an RTX2080TI GPU, and 32GB RAM.

\subsubsection{Evaluation Metrics}

\textbf{Evaluating Classification Performance.}
We adopt Micro-F1 and Macro-F1 scores as evaluation metrics for the classification tasks.
Mathematically, they are defined as below:
\begin{align}
\label{eq:micro-f1}
{\rm MicroF1} &= \frac{2\sum_{i=1}^{c}{\rm TP}_i}{\sum_{i=1}^{c}\left(2{\rm TP}_i+{\rm FP}_i+{\rm FN}_i\right)}\\
\label{eq:macro-f1}
{\rm MacroF1} &= \frac{1}{c}\sum_{i=1}^{c}\frac{2{\rm TP}_i}{2{\rm TP}_i+{\rm FP}_i+{\rm FN}}
\end{align}
where $c$ is the total number of classes and ${\rm TP}_i$, ${\rm FP}_i$, and ${\rm FN}_i$ respectively are the true positive, false positive, and false negative of the predicted results for the $i$th class.
In multi-class classification, the Micro-F1 scores measure the overall classification accuracy regardless of the performance in classifying individual classes.
The Macro-F1 scores are the mean of class-wise F1 scores in which it is sensitive to the classification performance of minority classes while the Micro-F1 scores are not.

\textbf{Evaluating Regression Performance.}
We evaluate the regression performance by measuring the Root Mean Square Error (RMSE) between the predicted values and the ground truths.
Mathematically, they are defined as below.
\begin{equation}
{\rm RMSE}=\sqrt{\frac{\sum_{y\in \mathcal{S}_{test}}\left(y-\hat{y}\right)^2}{|\mathcal{S}_{test}|}}
\end{equation}
RMSE reveals the regression error between the truth value $y$ and the predicted one $\hat{y}$ in the test set $\mathcal{S}_{test}$.

\subsection{Experimental Results and Analysis}

We conduct experiments to validate the effectiveness and efficiency of our proposed TADGE and report the results.

\begin{figure*}[t]
	\centering
	\subfigure[Results in the Transaction dataset]{
		\label{subfig_temporal_link_prediction:Transaction}
		\includegraphics[width=0.3\textwidth]{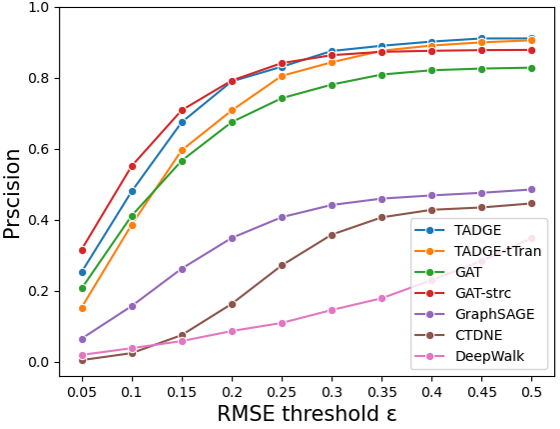}}
	\hspace{0.1in}
	\subfigure[Results in the Hyperlink dataset]{
		\label{subfig_temporal_link_prediction:Hyperlink}
		\includegraphics[width=0.3\textwidth]{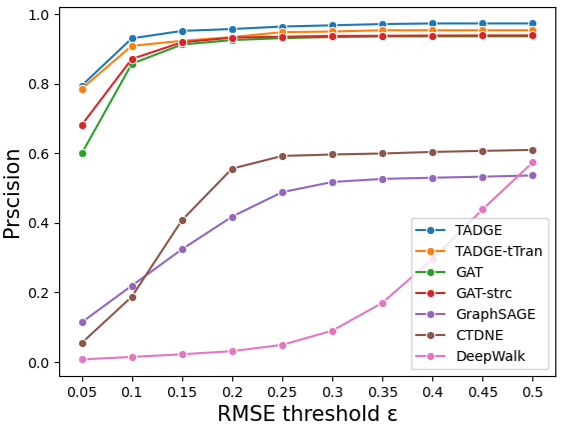}}
	\hspace{0.1in}
	\subfigure[Results in the Discussion dataset]{
		\label{subfig_temporal_link_prediction:Discussion} 
		\includegraphics[width=0.3\textwidth]{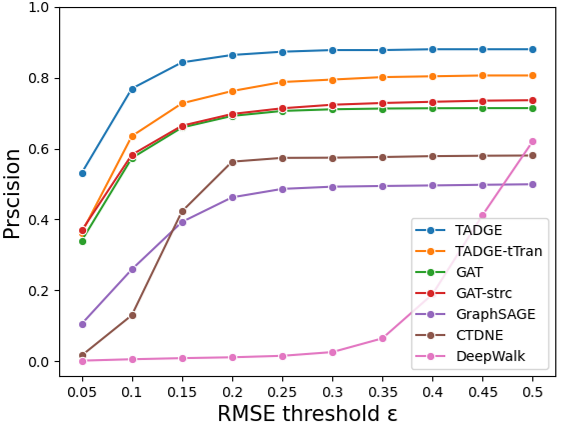}}
	\caption{Precision of the time-aware edge prediction with varying RMSE threshold $\varepsilon$.}
	\label{fig:temporal_link_prediction}
\end{figure*}

\subsubsection{ToE Prediction}

Given the embedding of vertex pairs, the ToE prediction is estimating the ToE of edge they formed, thus testing how effectively the learned vertex representations preserve the temporal dynamics.
For DeepWalk, CTDNE and GraphSAGE, which do not leverage ToE regression as a training task, we make use of the embeddings obtained in their training phases to train a regression model that employs $\mathcal{L}_{ToE}$ in Eq. (\ref{eq:task_ToE_reg}) as the objective function and adds an elastic net as the regularization.

The test results of ToE prediction are presented in the Table \ref{tab:result_ToE_Pred}.
The lower the RMSE, the more accurate the ToE prediction.
Our TADGE achieves the lowest RMSE and dramatically outperforms all baseline methods in three datasets, demonstrating that the temporal dynamic has been well preserved by the TADGE.
Comparing to those merely embedding the dynamic edge formation, i.e., GAT and TADGE-tTran, the ToE prediction error drops a lot when they further embed the asynchronous evolutions of local structures, thus validating the benefits of preserving asynchronous structural evolutions in ToE prediction.

\begin{table}[t]
	\centering
	\caption{RMSE of ToE prediction}
	\label{tab:result_ToE_Pred}
	\begin{tabular}{cccc}
		\toprule
		& Transaction & Hyperlink & Discussion\\
		\midrule
		DeepWalk & 0.5586 & 0.4667 & 0.4602 \\
		CTDNE & 0.4397 & 0.4582 & 0.4718 \\
		GraphSAGE & 0.4465 & 0.4464 & 0.4748 \\
		GAT & 0.4304 & 0.4125 & 0.3394 \\
		GAT-strc & 0.4025 & 0.4099 & 0.3317 \\
		TADGE-tTran & 0.3959 & 0.4073 & 0.3298 \\
		\textbf{TADGE} & \textbf{0.3795} & \textbf{0.4004} & \textbf{0.3170}\\
		\bottomrule
	\end{tabular}
\end{table}

\subsubsection{Static Edge Prediction}

The objective of static edge prediction is to determine whether a pair of vertices will form an edge in future timestamps when giving their current representations.
This task ignores the ToV, which is widely adopted by existing work to validate the performance of embedding algorithms in preserving the linkage structures of the graph.
We employ the cosine distance to measure the similarity of vertices' representation and predict whether they form an edge by a sigmoid function.

There are two ways to conduct the experiments of edge prediction.
One is to treat it as a binary classification problem, predicting whether an edge exists with the learned representations of two vertices.
Precision and recall are usually employed as evaluation metrics.
This protocol focuses more on evaluating how many edges the model can correctly predict while neglecting each vertex's prediction performance.
Therefore, even when the model fails in predicting edges for the minority of vertices, it will not affect the overall precision and recall much.
Another experiment protocol treats the edge prediction as a multi-class classification problem, identifying whether any given vertex will form an edge with all the rest vertices.
Micro-F1 and Macro-F1 scores are usually adopted to measure the prediction performance. 
The Micro-F1 score measures the overall prediction accuracy regardless of the individual performance of each vertex, which achieves the same goal as the former experiment protocol.
Meanwhile, the Macro-F1 score indicates the vertex-wise average performance regardless of how often they appear in the graph.
Therefore, this experiment protocol can reveal the performance of models in edge prediction much more comprehensively than the former one.
In this study, we adopt the latter one for edge prediction and the results are summarized in Table \ref{tab:result_static_linkPred}.

\begin{table}[t]
	\centering
	\caption{Results of Static Edge Prediction}
	\label{tab:result_static_linkPred}
	\resizebox{0.49\textwidth}{!}{
	\begin{tabular}{ccccccc}
		\toprule
		\multirow{2}*{  } & \multicolumn{2}{c}{Transaction} & \multicolumn{2}{c}{Hyperlink} & \multicolumn{2}{c}{Discussion}\\
		& Micro-F1 & Macro-F1 & Micro-F1 & Macro-F1 & Micro-F1 & Macro-F1\\
		\midrule
		DeepWalk & 0.8783 & 0.7012 & 0.8587 & 0.6476 & 0.8486 & 0.5115\\
		CTDNE & 0.5177 & 0.4103 & 0.7253 & 0.4527 & 0.6897 & 0.3839\\
		GraphSAGE & 0.6132 & 0.5157 & 0.7114 & 0.3732 & 0.6125 & 0.3788\\
		GAT & 0.8300 & 0.8033 & 0.9362 & 0.9071 & 0.7140 & 0.6794\\
		GAT-strc & 0.8787 & 0.8627 & 0.9389 & 0.9140 & 0.7377 & 0.7050\\
		TADGE-tTran & 0.9111 & 0.8864 & 0.9536 & 0.9328 & 0.8060 & 0.7721\\
		\textbf{TADGE} & \textbf{0.9133} & \textbf{0.8939} & \textbf{0.9732} & \textbf{0.9612} & \textbf{0.8799} & \textbf{0.8567}\\
		\bottomrule
	\end{tabular}}
\end{table}

Our TADGE achieved the highest Micro-F1 and Macro-F1 scores in all three datasets.
In our Time-aware Transformer, it first learns vertices' self-attentions within the local structure by an encoder.
Regarding this encoded vertex attentions as the context information, a decoder is then built to further embed the exact connections between the vertex pairs within the local structure, thus better preserving the dynamic edge formation and resulting in higher F1 scores than other baseline methods.
Besides, embedding the asynchronous structural evolutions further boosts the effectiveness of preserving the dynamic edge formation.
Consequently, our TADGE dramatically outperforms all baseline methods, especially in the Discussion dataset which is with the largest scale and highly dynamic connections.

\subsubsection{Time-aware Edge Prediction}

Time-aware edge prediction is a specific application for dynamic graph embedding abstracted from many data mining applications such as forecasting future crowd flow and recommending items at varying time intervals.
It simultaneously performs static edge prediction and ToE prediction.
An edge is correctly predicted if and only if true positive results are obtained in static edge prediction and the RMSE of ToE estimation is smaller than a threshold $\varepsilon$.
In order to expose the effect of ToE prediction error on the performance of the time-aware edge prediction, we test the threshold $\varepsilon$ from $0.05$ to $0.5$.
Because the performance of ToE prediction will not affect the results of negative edge samples in the time-aware edge prediction, we conduct the experiment on the positive edge samples in the test set and report the overall precision in Fig. \ref{fig:temporal_link_prediction}.

Our TADGE performs the best in the Hyperlink and the Discussion when $\varepsilon \ge 0.05$.
This shows that the RMSE of ToE prediction for most edges predicted by our TADGE in these datasets is less than $0.05$.
Although the TADGE performs slightly worse than the GAT-strc in the Transaction with small $\varepsilon$, it becomes the best of all when $\varepsilon>0.25$.
The results of DeepWalk show totally different trend due to its poor performance in ToE prediction.
The performance superiority of TADGE against DeepWalk, CTDNE and GraphSAGE demonstrates that temporal information carried by the ToV and ToE is a key aspect of preserving the dynamics connection between vertices.
Consequently, our TADGE preserves the spatial and temporal dynamics of the edge formation very well, therefore resulting in superior performance in the time-aware edge prediction.

\subsubsection{Vertex Classification}

\begin{table}[t]
	\centering
	\caption{Results of Vertex Classification}
	\label{tab:result_vertex_clf}
	\resizebox{0.45\textwidth}{!}{
		\begin{tabular}{ccccc}
			\toprule
			\multirow{2}*{  } & \multicolumn{2}{c}{Transaction} & \multicolumn{2}{c}{Hyperlink} \\
			& Micro-F1 & Macro-F1 & Micro-F1 & Macro-F1 \\
			\midrule
			DeepWalk & 0.5889 & 0.3721 & 0.7179 & 0.5185 \\
			CTDNE & 0.5969 & 0.3680 & 0.7119 & 0.4653 \\
			GraphSAGE & 0.6037 & 0.3711 & 0.7939 & 0.7361 \\
			GAT & 0.5952 & 0.4256 & 0.7156 & 0.6705 \\
			GAT-strc & 0.6173 & 0.4424 & 0.8067 & 0.7562 \\
			TADGE-tTran & 0.6042 & 0.4122 & 0.7596 & 0.7247 \\
			\textbf{TADGE} & \textbf{0.6302} & \textbf{0.4644} & \textbf{0.8140} & \textbf{0.7634} \\
			\bottomrule
	\end{tabular}}
\end{table}

Vertex classification aims to identify the unique labels of the vertices using their learned embeddings.
A support vector machine (SVM) with a Gaussian kernel is trained by using the embeddings obtained from the training set with the corresponding vertex labels.
After obtaining vertices' embeddings in the test set, we input them into the well-trained SVM and classify their labels.
It tests how well the embedding algorithm is in preserving the evolutionary patterns.
Since the Discussion dataset does not contain vertex labels, we compare the classification performance measured by the Micro-F1 and Macro-F1 scores in both Transaction and Hyperlink datasets.

The results are presented in Table \ref{tab:result_vertex_clf}. At the first glance, it is clear that our TADGE achieves the highest Micro-F1 and Macro-F1 scores in both datasets.
Remarkably, when neglecting the asynchronous structural evolving and merely embedding the pair-wise dynamic connection changes, the classification accuracy drops significantly.
Since TADGE and GAT-strc are trained by the same loss functions, the performance improvement of TADGE comes from the excellent preservation of dynamic connection changes by the Time-aware Transformer and super expression ability of our proposed attention mechanism.
Comparing to CTDNE and GraphSAGE, the superior performance of TADGE demonstrates that embedding the temporal information, i.e., ToV and ToE, makes the embeddings more discriminative.
In summary, our TADGE is capable of well preserving the asynchronous structural evolution patterns of vertices, thus leading to better vertex classification performance.

\subsubsection{Parameter Sensitivity Analysis}

We investigate the performance fluctuations of TADGE with varied hyper-parameters.
In particular, we study the sensitivity of TADGE to the embedding dimension $k$ and the number of blocks $N$ and heads $\bar{o}$ in static edge prediction and ToE prediction using the Hyperlink dataset.
We vary the value of one hyper-parameter while fixing the others.

\begin{figure*}[t]
	\centering
	\subfigure[Micro-F1 scores of static edge prediction]{
		\label{subfig_test_block_head:Micro-f1_linkPred}
		\includegraphics[width=0.3\textwidth]{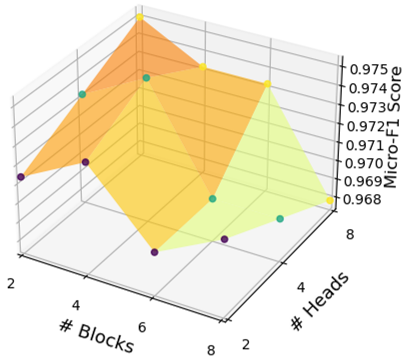}}
	\hspace{0.1in}
	\subfigure[Macro-F1 scores of static edge prediction]{
		\label{subfig_test_block_head:Macro-f1_linkPred}
		\includegraphics[width=0.3\textwidth]{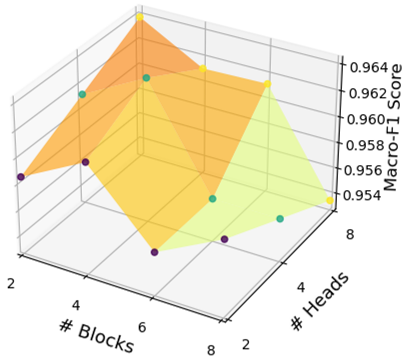}}
	\hspace{0.1in}
	\subfigure[RMSE of ToE prediction]{
		\label{subfig_test_block_head:ToE_pred}
		\includegraphics[width=0.3\textwidth]{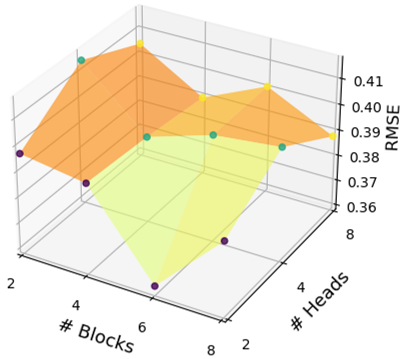}}
	\caption{Testing the hyper-parameters with varying blocks $N$ and heads $\bar{o}$ in the Hyperlink.}
	\label{fig:test_block_head}
\end{figure*}

\textbf{Impact of the number of blocks $N$ and heads $\bar{o}$.}
The number of blocks and heads are the key parameters affecting the expression ability of the self-attention mechanism in the TADGE to learn the asynchronous structural evolutions.
We test the combination of $N\in\{2,4,6,8\}$ and $\bar{o}\in\{2,4,8\}$ to evaluate their impacts on the performance fluctuations of TADGE.
As the results in Fig. \ref{fig:test_block_head}, in the static edge prediction, the increasing number of blocks has a negative impact on both Micro-F1 and Macro-F1 scores while, in general, more heads the better performance.
Similar trends are observed in ToE prediction.
In summary, TADGE prefers more heads but fewer blocks so as to better embed the asynchronous structural evolutions in the dynamic graph.

\begin{figure}[t]
	\centering
	\subfigure[Results in the self-identification of vertex and static edge prediction]{
		\label{subfig_test_k:seld-iden_link_pred}
		\includegraphics[width=0.235\textwidth]{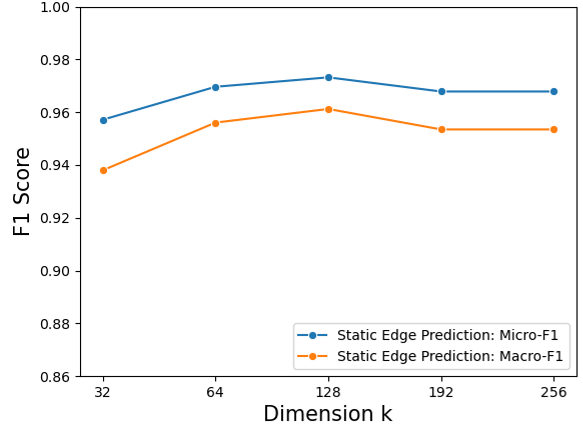}}
	\subfigure[Results in the ToE prediction]{
		\label{subfig_test_k:ToE_pred}
		\includegraphics[width=0.235\textwidth]{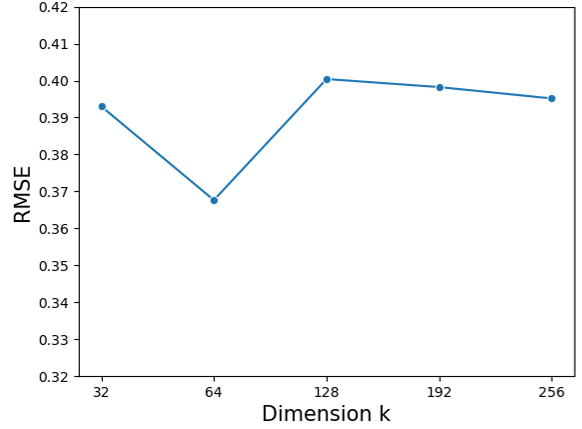}}
	\caption{Testing the dimension of representation $k$ in the Hyperlink.}
	\label{fig:test_k}
\end{figure}

\textbf{Impact of $k$.}
The embedding dimension $k$ is an important hyper-parameter affecting the expressiveness of TADGE.
We exterminate $k$ in $\{32,64,128,192,256\}$.
As shown in Fig. \ref{fig:test_k}, in static edge prediction, both Micro-F1 and Macro-F1 scores increase at the beginning, indicating that the expressive power of the embedding grows as $k$ increases
After reaching the best F1 score when $k=128$, the classification performance slightly drops afterward.
In ToE prediction, TADGE gets the smallest RMSE when $k=64$, and the prediction errors with the other $k$ values are close.
Hence, TADGE is not sensitive to $k$ in ToE prediction.

\subsubsection{Convergence and Training Efficiency Analysis}

\begin{figure*}[ht]
	\centering
	\subfigure[Training loss]{
		\label{subfig_convergence_efficiency:loss}
		\includegraphics[width=0.23\textwidth]{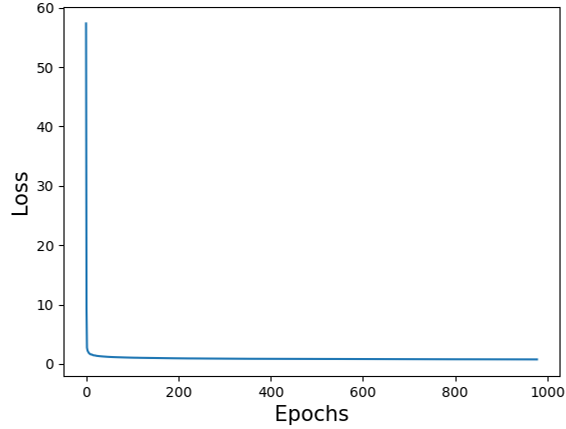}}
	\hspace{0.07in}
	\subfigure[Training F1 scores]{
			\label{subfig_convergence_efficiency:f1_loc_linkPred}
			\includegraphics[width=0.23\textwidth]{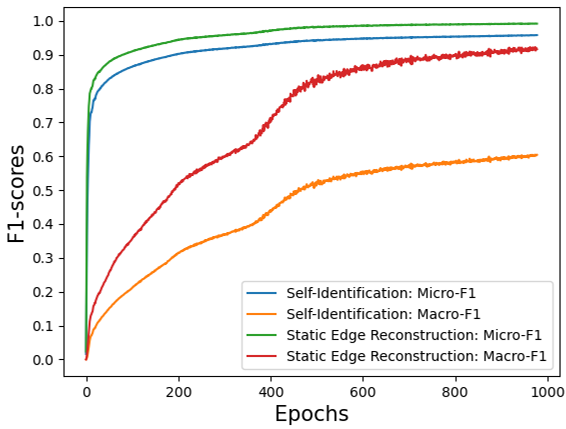}}
		\hspace{0.07in}
		\subfigure[Training RMSE]{
				\label{subfig_convergence_efficiency:RMSE_ToE_strcEvoTime}
				\includegraphics[width=0.23\textwidth]{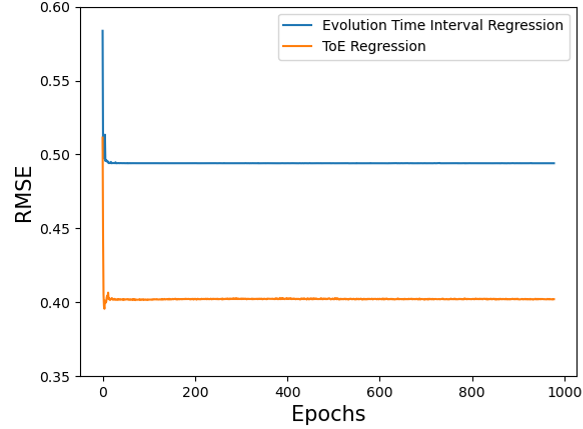}}
			\hspace{0.07in}
			\subfigure[Training time with varying $k$]{
				\label{subfig_convergence_efficiency:training_time_k}
				\includegraphics[width=0.23\textwidth]{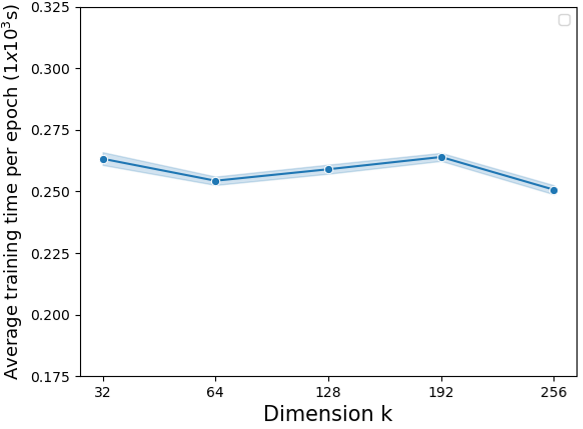}}
			\caption{Convergence and training efficiency of TADGE in the Hyperlink. The shadowed area in (d) shows the variance of training time.}
			\label{fig:convergence_efficiency}
\end{figure*}

We demonstrate the convergence and training efficiency of TADGE in the Hyperlink datasets.
In Fig. \ref{subfig_convergence_efficiency:loss}, we observe that training loss drops quickly and converges within a hundred epochs.
The overall training accuracy in both edge reconstruction and self-identification of vertices quickly converges to over $0.9$ Micro-f1 score as shown in Fig. \ref{subfig_convergence_efficiency:f1_loc_linkPred}.
Although the loss changes slightly, the Macro-f1 scores in both tasks gradually increase and eventually reach the convergence, which is consistent with the conclusion drawn in \cite{vaswani2017attention}.
Because the appearing times of vertices in a dynamic graph follow the long-tailed distribution, there are a number of vertices having very few connections to others.
They fail to provide enough linkage information for the embedding algorithms to learn.
Consequently, in the self-identification tasks, the TADGE converges to very high Micro-F1 scores but with relatively low Macro-F1 scores.
Better dealing with this minority of vertices will be a direction for further extending this study.
In both ToE prediction and structural evolution time interval estimation, which are two regression tasks, the training RMSE converges within $20$ epochs as shown in Fig. \ref{subfig_convergence_efficiency:RMSE_ToE_strcEvoTime}, indicating the advantageous convergence speed of the TADGE in the regression tasks.

Fig. \ref{subfig_convergence_efficiency:training_time_k} shows the average running time of every epoch while training the TADGE with varying dimension $k$.
The shadow shows the variance of training time.
As $k$ increases, the running time of updating the trainable parameters of TADGE at each epoch fluctuates slightly.
This validates that the training time is not sensitive to the dimension of the representations.
Hence, we conclude that TADGE has very good convergence and training efficiency.

\subsubsection{Scalability Analysis}

\begin{figure}[t]
	\centering
	\includegraphics[width=0.3\textwidth]{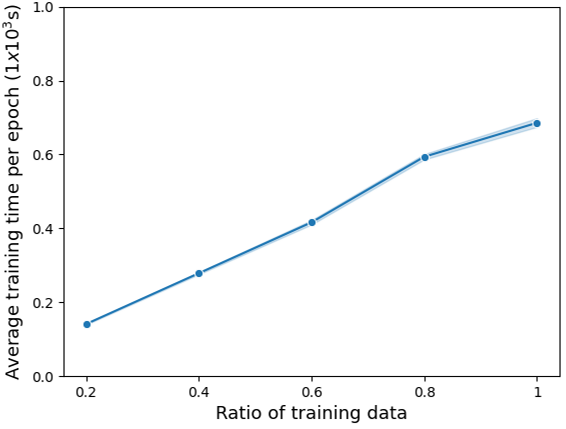}
	\caption{Training time of TADGE per epoch with varying data proportions.}
	\label{fig:scalability}
\end{figure}

We conduct a scalability test for the TADGE in the Discussion dataset which contains $194,085$ vertices and $1,443,339$ edges.
We vary the proportion of the training data in $\{0.2, 0.4, 0.6, 0.8, 1.0\}$ to train the TADGE and report the average training time of every epoch.
Ideally, the training time should increase linearly when we enlarge the scale of training data.
The growth of training time is shown in Fig. \ref{fig:scalability}.
The line indicates the average running time of updating the weights of TADGE at each epoch and the shadow shows the variance.
As the scale of training data gradually enlarges, the average training time per epoch grows linearly from $0.141\times10^3$ seconds to $0.686\times10^3$ seconds.
Consequently, we conclude that the TADGE has very good scalability for embedding very large-scale dynamic graphs.

\subsubsection{Analysis of Initialization Methods for Training TADGE}

\begin{figure}[t]
	\centering
	\includegraphics[width=0.3\textwidth]{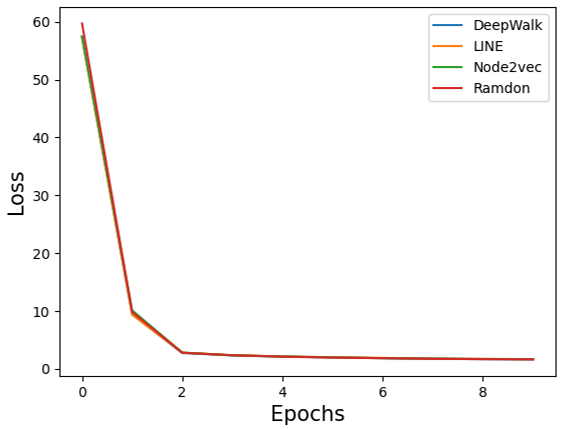}
	\caption{Training loss of TADGE using different initialization methods.}
	\label{fig:loss_initialization}
\end{figure}

To evaluate the impact of initialization approaches on the effectiveness and training efficiency of TADGE, we respectively employ DeepWalk~\cite{perozzi2014deepwalk}, Node2vec~\cite{node2vec}, and LINE~\cite{tang2015line} to initialize the representation $R_v$ and train TADGE in the Hyperlink dataset.
All these approaches are trained on a static graph constructed from the edges in the training set.
We also adopt a random initialization following the normal distribution as one of the baselines.
The training loss is shown in Fig. \ref{fig:loss_initialization}.
Although the initial loss of using static graph embedding approaches as the initialization is smaller than that of using a random one, the training loss shows almost the same trend with a very slight difference since the first epoch.
This confirms that the outstanding convergence speed of TADGE is because of the approach itself rather than the initialization approach.
Table \ref{tab:result_initialization} reports the results of TADGE in the ToE prediction, static edge prediction, and vertex classification when initializing the training using different approaches.
The results are close to each other, demonstrating that our TADGE is not sensitive to the initialization approaches.

\begin{table}[t]
	\centering
	\caption{Results of TADGE with Varying Initialization Methods in the Hyperlink}
	\label{tab:result_initialization}
	\resizebox{0.48\textwidth}{!}{
		\begin{tabular}{ccccccc}
			\toprule
			\multirow{2}*{} & \multicolumn{2}{c}{Static Edge Prediction} & \multicolumn{2}{c}{Vertex Classification} & \multicolumn{1}{c}{ToE Prediction} \\
			 & Micro-F1 & Macro-F1 & Micro-F1 & Macro-F1 & RMSE \\
			\midrule
			Random & 0.9649 & 0.9511 & 0.8276 & 0.7550 & 0.4029\\
			DeepWalk & 0.9732 & 0.9612 & 0.8140 & 0.7634 & 0.4004\\
			Node2vec & 0.9666 & 0.9534 & 0.8123 & 0.7598 & 0.3984\\
			LINE & 0.9631 & 0.9485 & 0.8246 & 0.7703 & 0.4007\\
			\bottomrule
	\end{tabular}}
\end{table}

\section{Related Work}

The main issue in dynamic graph embedding is capturing and encoding the dynamic evolving nature of vertices and edges.
Modeling the dynamic graph in a proper manner is the foundation as it captures the dynamics of vertices and edges for later embedding.
Existing approaches model the dynamic graph as either a snapshot graph sequence (SGS)~)~\cite{zhu2016scalable}\cite{goyal2019dyngraph2vec}\cite{zhu2018high}\cite{zhou2018dynamic}\cite{velickovic2018graph}\cite{manessi2020dynamic}\cite{xiong2019dyngraphgan} or neighborhood formation sequences (NFS) sampled from the temporal random walk~\cite{nguyen2018continuous}\cite{fathy2020temporalgat}\cite{chang2020continuous}\cite{zuo2018embedding}.
These approaches merely capture the synchronous structural evolutions and their limitations have been discussed in Section \ref{sec:Introduction}.
Our dynamic graph model captures not only the dynamic connection changes among vertices but also the asynchronous evolution starting time and duration for embedding.

Given the above dynamic graph models, the embedding algorithm aims to encode the captured evolution patterns as representations of vertices or edges.
TNE \cite{zhu2016scalable} is a pioneer work that embeds the structural difference in SGS.
A series of similar approaches are continuously published, such as DHPE \cite{zhu2018high}, TMF \cite{yu2017temporally}, and DynGraph2Vec \cite{goyal2020dyngraph2vec}.
Instead of measuring the overall difference between snapshots, DynamicTriad \cite{zhou2018dynamic} embeds the triad formation process among vertices when a graph evolves.
GraphSAGE \cite{hamilton2017graphSAGE} learns the structure by a graph neural network, leveraging neighborhood information to generate representations for the new coming vertices.
Du, et al.~\cite{du2018dynamic} proposed an generic framework to extend skip-gram-based static graph embedding methods to update vertices' representations when the graph evolves and generate embeddings for new vertices.
PME\cite{chen2018pme} and MGCN\cite{chen2020multi} borrowed the similar ideas of embedding SGS, learning the correlation among multiple sub-graphs for highly accurate link prediction.

Learning the sequential edge formation was originally proposed in \cite{nguyen2018continuous} which embedding continuous-time dynamic graph.
HTNE\cite{zuo2018embedding} samples NFS by a multivariate Hawkes process and employ an attention network to embed sequential edge formation.
Following this idea, a series works have been proposed to learn the sequential patterns from edge sequences \cite{fathy2020temporalgat}\cite{chang2020continuous} by using the graph attention \cite{velickovic2018graph}\cite{sankar2020dysat}\cite{fathy2020temporalgat}, generative adversarial networks \cite{xiong2019dyngraphgan}\cite{zhou2020data}.
TMER\cite{chen2021temporal} proposed temporal meta-path to obtain sequential patterns for recommendation.
Although none of them embed the asynchronous structural evolutions, they inspire us to design the TADGE for jointly embedding the pair-wise connections and the local structures.

There are a few works focusing on temporal network embedding.
M$^2$DNE \cite{lu2019temporal} embeds the temporal edge formation process and the evolving scale of the graph.
It introduces a time decay function while calculating the attention value between connected vertices, which treats the temporal information as an edge attribute for embedding.
EPNE \cite{wang2020epne} embeds the periodic connection changes among vertices by causal convolutions.
TCDGE \cite{yang2021time} co-trains a liner regressor to embed ToE while learning the connection difference between consecutive snapshots.
Although references \cite{rossi2020temporal} and \cite{kumar2019predicting} claim themselves as the temporal network embedding, they degenerate temporal networks as NFS before embedding, thus merely preserving sequential structural evolutions without temporal information.
None of above mentioned works deals with the dynamic ToV and ToE at the same time.
Hence, they fail to embed the asynchronous structural evolutions but we fill this research gap in this paper.

\section{Conclusions}
We generically model a dynamic graph as a set of temporal edges, appending the respective joining time of vertices (ToV) and timespan of edges (ToE).
A time-centrality-biased temporal random walk is proposed to sample the dynamic graph as a set of temporal edge sequences for capturing the asynchronous structural evolutions.
A TADGE model containing a Time-aware Transformer and a structural embedding model is then proposed to simultaneously embed the dynamic connection changes with ToE and the asynchronous evolution starting time of every local structure.
The experimental results show that our TADGE achieves significant performance improvement over the state-of-the-art approaches in various data mining tasks, thus validating the effectiveness of TADGE to embed the asynchronous structural evolutions.
Besides, TADGE is very efficient and scalable when handling large-scale dynamic graphs.